# Gradual drift detection in process models using conformance metrics

Víctor Gallego-Fontenla[✉], Juan Vidal, Manuel Lama

Centro Singular de Investigación en Tecnoloxías Intelixentes (CiTIUS)
Universidade de Santiago de Compostela (USC), Galicia, Spain

{victorjose.gallego, juan.vidal, manuel.lama}@usc.es

**Changes, planned or unexpected, are common during the execution of real-life processes. Detecting these changes is a must for optimizing the performance of organizations running such processes. Most of the algorithms present in the *state-of-the-art* focus on the detection of sudden changes, leaving aside other types of changes. This paper focuses on the automatic detection of gradual drifts, a special type of change, in which the cases of two models overlap during a period of time. The proposed algorithm relies on conformance checking metrics to carry out the automatic detection of the changes, performing also a fully automatic classification of these changes into sudden or gradual. The approach has been validated with a synthetic dataset consisting of 120 logs with different distributions of changes, getting better results in terms of detection and classification accuracy, delay and change region overlapping than the main *state-of-the-art* algorithms.**

**Acknowledgements:** The work from Víctor J. Gallego was supported by the Spanish Ministerio de Educación, Cultura y Deporte (grant FPU17/05138, co-funded by the European Regional Development Fund – ERDF program); the Galician Consellería de Educación, Universidade e Formación Profesional (accreditation 2019-2022, ED431G-2019/04) and the European Regional Development Fund (ERDF). This paper was also supported by the Spanish Ministerio de Ciencia e Innovación under projects PDC2021-121072-C21 and PID2020-112623GB-I00.

## 1   Introduction

While the day-to-day operations of organizations normally remain unchanged over time, certain processes are susceptible to change due to internal or external factors. Regulatory shifts, changes in availability of different resources or new consumption patterns, for example, can influence the structure of the processes that shape these operations. At best, these changes will be carried out intentionally, so the organization will be aware of their presence. But usually, these changes arise without prior warning and may be overlooked. It is in these cases when it becomes important to have methods that allow organizations to detect these changes in order to explicitly adapt their processes. Otherwise the organization might take decisions on the basis of outdated knowledge and, consequently, get unexpected results.

| Source | Trace | Source | Trace | Source | Trace | Source | Trace |
|---|---|---|---|---|---|---|---|
| $M_1$ | A B C D | $M_1$ | A B C D | $M_1$ | A B C D | $M_1$ | A B C D |
| $M_1$ | A B C D | $M_1$ | A B C D | $M_1$ | A B C D | $M_1$ | A B C D |
| $M_1$ | A B C D | $M_1$ | A B C D | $M_1$ | A B C D | $M_1$ | A B C D |
| $M_1$ | A B C D | $M_2$ | A B D C | $M_1$ | A B C D | $M_2$ | A B D C |
| $M_1$ | A B C D | $M_1$ | A B C D | $M_2$ | A B D C | $M_2$ | A B D C |
| $M_1$ | A B C D | $M_1$ | A B C D | $M_2$ | A B D C | $M_2$ | A B D C |
| $M_2$ | A B D C | $M_2$ | A B D C | $M_2$ | A B D C | $M_3$ | A B D C E |
| $M_2$ | A B D C | $M_2$ | A B D C | $M_2$ | A B D C | $M_3$ | A B D C E |
| $M_2$ | A B D C | $M_1$ | A B C D | $M_1$ | A B C D | $M_3$ | A B D C E |
| $M_2$ | A B D C | $M_2$ | A B D C | $M_1$ | A B C D | $M_4$ | A D C E |
| $M_2$ | A B D C | $M_2$ | A B D C | $M_1$ | A B C D | $M_4$ | A D C E |
| $M_2$ | A B D C | $M_2$ | A B D C | $M_1$ | A B C D | $M_4$ | A D C E |
| (a) Sudden drift | | (b) Gradual drift | | (c) Recurrent drift | | (d) Incremental drift | |

Figure 1: Types of concept drift based on their occurrence over time.

These changes, also called *drifts*, are categorized into four types, depending on their temporal distribution (Gama et al. 2014): sudden drifts (Figure 1a), when they occur at an exact moment in time; gradual drifts (Figure 1b), when the change is prolonged over time, with the original and the modified behavior overlapping for a period of time; recurrent drifts (Figure 1c), when the changes repeat from time to time; and incremental drifts (Figure 1d), when successive changes with an intermediate behavior repeat until the model stabilizes. Despite being a widely explored field in descriptive analysis, it should be noted that dealing with concept drift in process mining faces an additional challenge: the data —execution traces from a process— may contain partial information about the process structure that can only be correctly understood when looking at the whole set of traces. For example, the presence of a choice structure can only be extracted by observing several traces, since the information present in a single trace will only represent a specific path. This can happen also with other control structures, such as loops or concurrent blocks.

This paper focus on the detection of gradual drifts, in which two versions of the process coexist for a certain period of time until one of them completely replaces the other. This type of change is challenging in process mining, since whenever a change is made to the process, traces already in execution can continue with the previous version of the model, while new traces will be generated by the already modified model, so both versions will coexist for a certain period of time. Thus, proposals that deal with gradual drift in process mining must ideally have a series of desirable features:

*F1. Automatically distinguish between gradual and sudden drifts*, so it is not know *a priori* which kind of changes might be found in the log. Note that there is no possible confusion between gradual drifts and the other two types of drifts, since in recurrent drifts we deal with sudden or gradual drifts that are periodically repeated, while incremental drifts are a squence of sudden or gradual drifts that constitute a single change.

*F2.* Deal with different *distributions of traces* during the occurrence of gradual changes such as linear, exponential, random or constant.

*F3.* Detect gradual changes with a *low delay*, so a change can be precisely localized in time.

*F4.* Have a *high accuracy*, detecting all the changes minimizing the *false positives* and *false negatives*.

Several authors have proposed a variety of approaches to the detection of sudden concept drift in process mining (Carmona and Gavaldà 2012; Junior et al. 2018; Maggi et al. 2013; Bose et al. 2014; Zheng et al. 2017; Gallego-Fontenla et al. 2021). In contrast, as a recent *state-of-the-art* survey on concept drift adoption (Sato et al. 2021) shows that only a few proposals allow automatic detection and classification of gradual changes, and their results still provide room for improvement. These proposals follow three main approaches for performing the drift detection: (i) extracting the trace features to abstract its behavior and then detect changes in these features (Bose et al. 2014; Martjushev et al. 2015; Luengo and Sepúlveda 2011; Li et al. 2017); (ii) generating process models and compute its features to detect changes (Tavares et al. 2019; Stertz and Rinderle-Ma 2018; Yeshchenko et al. 2019); or (iii) a combination of both approaches (Maaradji et al. 2015, 2017). However, these approaches present some drawbacks. Some of them address either sudden or gradual drifts, but not both at the same time (Bose et al. 2014; Martjushev et al. 2015), while other ones do not even distinguish between sudden and gradual drifts (Luengo and Sepúlveda 2011; Li et al. 2017; Tavares et al. 2019). In other cases, only a subset of the possible change patterns are detected (Stertz and Rinderle-Ma 2018), or the detection must be manually confirmed by the user (Yeshchenko et al. 2019). Furthermore, some approaches support only one distribution of traces during the changes, usually the linear distribution (Bose et al. 2014; Maaradji et al. 2015, 2017). Finally, most of these approaches have been only tested with 5 or less logs (Stertz and Rinderle-Ma 2018; Bose et al. 2014; Martjushev et al. 2015; Luengo and Sepúlveda 2011; Li et al. 2017; Yeshchenko et al. 2019) —only Tavares et al. (2019) has used 18 logs in its validation— so it is difficult to assess the quality of each approach.

In this paper, we present *CRIER* (Conformance-based gRadual drIft detEction algoRithm), an algorithm that can detect gradual and sudden changes in event logs based on the change of conformance checking metrics —fitness and precision— over a sliding window. *CRIER* computes these metrics for the successive sliding windows from the starts until the end of the log. The evolution of these conformance metrics is evaluated using linear regressions and hypothesis testing, so that a potential change is detected when the metric values change significantly. In order to identify gradual drifts, detected change points are analyzed. If the behaviour contained between two consecutive points is a combination of the behaviour present before the first point —detected as a change in fitness— and the behaviour after the second one —detected as a change in precision—, then the change is classified as gradual. *CRIER* has been tested using a dataset with 120 synthetic event logs with different traces distributions. The results show that our approach improves significantly the scores obtained by the main approaches of the *state-of-the-art*, making possible the treatment of sudden and gradual drifts in logs without any user intervention.

The remainder of the paper is structured as follows. In Section 2 we present the main approaches to gradual concept drift detection in process mining from the *state-of-the-art*. In Section 3 we define a set of concepts that define the framework of the approach. In Section 4 we present the formal proof of the hypothesis behind the proposed algorithm. In Section 5 we detail how *CRIER* is able to detect both sudden and gradual changes automatically. In Section 6 we present the validation of our approach and how it outperforms the main algorithms from the literature. Finally, in Section 7 we present our conclusions and outline our future work.

## 2 Related Work

Concept drift is a traditional problem in descriptive analysis that studies the change over time of a target variable in order to minimise the error of a prediction. Although, in general, it is a widely researched and discussed topic in machine learning, it has received little attention in the field of process mining. However, concept drift is recognised as a significant part of process mining (van der Aalst et al. 2011), and essential to deal with the changing nature of business processes. A recent survey on concept drift in process mining (Sato et al. 2021) identifies a total of 36 scientific publications and 22 different approaches of which only 5 can handle gradual changes.

A common approach to gradual change detection in process mining is the transformation of logs into time series by extracting different features from the traces. These features make it possible to obtain useful information contained in the traces, simplifying the data to be processed. Based on this idea, Bose et al. (2014) present an approach that computes features from follow/precede relations and use a statistical hypothesis test to check if there are changes in the features over time, specifically over two consecutive fixed size windows. Examples of these features are: for each activity of the log, the number of activities that always, sometimes and never follow/precede a given activity; for each pair of activities $\alpha_1$ and $\alpha_2$ of the log and for each trace, the number of sequences of size *n* that start with $\alpha_1$ and contain $\alpha_2$; for each pair of activities $\alpha_1$ and $\alpha_2$ of the log and for each trace, the significance of $\alpha_1$ following/preceding $\alpha_2$ with a maximum distance of *n*; etc. The main drawback of this approach is that it does not distinguish between sudden and gradual changes and it deals only with linear distributions for gradual changes. Additionally, it requires the user to have an advanced knowledge of the process, including which features to select, the activities that can change or the statistical test to be used in order to process the log. Furthermore, the approach is not extensively tested, being validated only with two synthetic logs and a real one, without providing any quantitative assessment of the results, although they seem to be promising. An extension to this approach is presented by Martjushev et al. (2015), where the authors propose the use of non-consecutive windows, leaving a gap between them, in order to increase the difference between features in gradual environments in order to reduce the false negatives. In this extension, the restriction of gradual drifts distribution is removed, so the algorithm can detect also changes that are not linear. Yet, the approach fails to provide a solution able to identify if a change is sudden or gradual, requires even more knowledge from the user to set other parameters —as the size of the gap between the windows—, and still does not provide quantitative metrics in the validation, which is performed only over 2 logs.

In Li et al. (2017) the log is transformed into a time series using a feature called *r-measure* that compares the relation matrices between two consecutive windows. Then, the algorithm looks for outliers in these time series, and determines that a change exists when one is detected. As the previous proposals, the main drawback of this approach is that it does not distinguish sudden and gradual drifts and it is prone to mix up changes with outliers —abnormal executions—. Moreover, it is also not extensively tested, being evaluated against 4 synthetic logs, where they obtain very promising results in terms of $F_{score}$, with values between 0.6 and 0.95 depending on the used parameters.

Luengo and Sepúlveda (2011) propose the use of an agglomerative hierarchical clustering over the traces to detect changes. For the clustering, traces are transformed to feature vectors that abstract the trace behaviour. Specifically, the maximal repeat (Bose and van der Aalst 2009) and the starting timestamp of the trace are used as features. Once the clusters are generated, they define a change as the point in time where one cluster ends and a new one starts. However, this method is very dependent on the number of clusters, which should be fixed by the user and equal to the number of changes present in the log. Regarding the experimentation, the approach is only validated using 3 logs, with an accuracy —calculated as the sum of true positives and true negatives, and divided by the total number of traces— between 57% and 100%.

Stertz and Rinderle-Ma (2018) propose an algorithm that deals with a stream of events, allowing the detection of changes in incomplete traces. Using a window of a given size, different versions of the process model are discovered —the *process history*— and the fitness is computed against the last mined model. In this approach, a change is present when the fitness of a trace is below a threshold. Thus, when a trace does not fit the last discovered model, a new one is mined using only the unfitting traces. To prevent false positives, each model also receives an score, and a model is considered to change only when that score is over a threshold. Finally, changes are classified in sudden, gradual, recurring and incremental using both two new thresholds and the process history. The main drawback of this approach is that it requires the user to have a deep knowledge of the problem in order to tune the hyperparameters used in the detection —the window size and 4 different thresholds—. Furthermore, the algorithm can not detect all structural change patterns, e.g., the transformation of an optional parallel structure to an exclusive choice where the order of some activities is enforced. Also, the approach has not been extensively tested and no quantitative measures are provided about the goodness of the results.

Yeshchenko et al. (2019) propose a method that is also based on the idea of using the process model for detecting changes. This approach splits the log in *n* windows of size *m*. Then, a set of declarative rules is extracted from the complete log and their confidence is computed with respect to each window, obtaining a multivalued time series. Finally, a traditional concept drift detection algorithm —namely, the PELT algorithm— in combination with a hierarchical clustering technique is used to detect changes in the resulting time series. The main drawback with this approach is that the classification of changes in sudden or gradual is not automatic, but it must be performed visually by the user. Regarding the validation, the algorithm shows promising results for the detection, but it is only tested with 4 synthetic and 2 real logs.

Another very interesting proposal, halfway between the extraction of features and the use of a process model, is the one presented by Maaradji et al. (2015, 2017). In this approach, the behavior of the traces is abstracted using *partial-ordered-runs* —i.e., a graph representation of the traces—. Once the behavior is abstracted, two consecutive sliding windows are used and, by means of a statistical test, it is checked if the content of these windows is significantly different. Once the changes are detected, a classification is performed to guess if they represent a sudden or a gradual drift. In this classification, a statistical test checks if the combination of traces between two consecutive changes represents a linear combination of the traces before and after the first and second changes, respectively. This is one of the most thoroughly validated proposals, tested with multiple logs and evaluated using well-known metrics, as $F_{score}$ and *delay*. The main drawback of this approach is that it can be very sensitive to changes in the frequencies of the relations, which may lead to the detection of false positives. Also, the algorithm can only detect gradual changes that are due to a linear distribution of the traces between two models, which may not be the case in real logs.

Finally, Tavares et al. (2019) propose an online approach where they use a clustering approach over graph distances. The approach takes as input an event stream, and updates the corresponding trace graph every time a new event is received. When they have enough traces, a process model is discovered by generating two weighted graphs: in one of them, the weight of the arcs represents the frequency of the transitions between activities; and in the other one, the average time between activities is the arc weight. Then, a distance between the trace graph and these two weighted graphs is computed, and these distances are grouped using a online density-based clustering algorithm —namely, DenStream—. Regarding the validation, the approach is tested using 18 synthetic logs, but the only metric provided is the number of changes detected. The main drawback of this method is that it mixes control flow and behavioral changes. Also, it requires a lot of hyperparameters to be tuned by the user in order to obtain good results. Finally, as many of the presented proposals, it supports the detection over logs with gradual changes, but does not distinguish sudden from gradual changes.

As a summary, there is no approach that provides a fully automated solution for detecting sudden and gradual changes, supporting all types of trace distributions and with a exhaustive validaton that demostrates its high accuracy and low delay.

## 3 Preliminaries

In this section, we introduce some terms necessary to understand the concept drift detection problem and our approach. The proposed method takes an event log as input and tries to detect changes in the observed behaviour.

**Definition 1 (Event).** Given the set of activities $\mathcal{A}$ that conform a process, an event $\varepsilon$ can be defined as the execution of an activity $\alpha \in \mathcal{A}$ at a given instant $t$ in the context of a process instance $c$. The activity name $\varepsilon.a$, the timestamp $\varepsilon.t$ and the process instance identifier —often referred to as *case*— $\varepsilon.c$ are the only mandatory attributes of an event, which can also have other generic attributes, such as the resources that perform the activities, or domain-specific attributes, understood as variables whose values are modified in the activity execution.

**Definition 2 (Trace).** Given the full set of events $\mathcal{E}$ recorded from the execution of a process, a trace $\tau$ can be defined as the ordered sequence of all the events belonging to the same process instance $\tau.c$, where the order is defined by the timestamp of the events.

$$\tau = \langle \varepsilon_0, \ldots, \varepsilon_n \rangle : \forall i \in [0, n), \forall j \in (i, n], \varepsilon_i, \varepsilon_j \in \mathcal{E} \rightarrow (\varepsilon_i.c = \varepsilon_j.c) \wedge (\varepsilon_i.t < \varepsilon_j.t).$$

| Case | Timestamp | Activity | Resource |
|------|-----------|----------|----------|
| #aaa | 01/10/2021 08:01 | Lock feature | Phoebe |
| #aaa | 01/10/2021 08:53 | Check restrictions | Phoebe |
| #aab | 01/10/2021 11:40 | Lock feature | Rachel |
| #aac | 01/10/2021 09:12 | Lock feature | Ross |
| #aac | 01/10/2021 09:33 | Interview customer | Ross |
| #aac | 01/10/2021 11:48 | Build part | Ross |
| #aab | 01/10/2021 11:49 | Check restrictions | Rachel |
| #aaa | 01/10/2021 08:57 | Build part | Phoebe |
| #aab | 01/10/2021 16:18 | Build part | Rachel |
| #aac | 01/10/2021 12:16 | Quality test | Monica |
| #aaa | 01/10/2021 13:45 | Integration test | Chandler |
| #aab | 01/10/2021 17:23 | Integration test | Joey |
| #aaa | 01/10/2021 13:37 | Quality test | Monica |
| #aac | 01/10/2021 16:22 | Integration test | Joey |
| #aab | 01/10/2021 17:35 | Quality test | Monica |

Figure 2: Example of a log with 15 events belonging to 3 diferent traces. It records the execution of 5 different activities by 6 resources.

We denote as $B_\tau$ the behaviour observed in a trace $\tau$, represented as an ordered sequence of activities:

$$B_\tau = \langle \alpha_0, \ldots, \alpha_n \rangle : \forall \epsilon_i \in \tau \rightarrow \alpha_i = \epsilon_i.a$$

**Definition 3 (Log).** A log can be defined as an sequence of traces $L = \langle \tau_0, \ldots, \tau_n \rangle$ where each trace represents a different process instance $\nexists \tau_i, \tau_j \in L : \tau_i.c = \tau_j.c$ and the order is given by the timestamp of the last event in each trace. We denote as $B_L = \{B_{\tau_0}, \ldots, B_{\tau_n}\}$ the behaviour observed in the log $L$. This behaviour is composed by the set of distinct behaviours captured in the traces of the log. Two different traces $\tau_i$ and $\tau_j$ such that $\tau_i.c \neq \tau_j.c$ can have the same behaviour $B_{\tau_i} = B_{\tau_j}$ if the ordered sequence of activities for $\tau_i$ is the same as for $\tau_j$. The size of $B_L$ will always be smaller than or, at most, equal to the size of the log $|B_L| \leq |L|$.

Figure 2 shows an example of an event log with 3 different traces, 15 events, 5 activities, and 6 resources. In this example, the behaviour observed in the log $B_L$ is:

$$B_L = \begin{cases} \langle \text{Lock feature, Check restrictions, Build part, Integration test, Quality test} \rangle \\ \langle \text{Lock feature, Interview customer, Build part, Quality test, Integration test} \rangle \end{cases}$$

being the first of the sequences the behaviour observed in cases #aaa and #aab and the second one the behaviour for the case #aac.

The relations and the dependencies between the different activities that conform a process can be represented using a *process model*, which can be represented with a directed graph that ideally describes the behaviour observed in the execution log. In this paper, we use Petri nets to represent the process models because they provide, simultaneously, a powerful mathematical formalism and a simple graphical representation of these models. However, the approach is not tied to this formalism, so any other language capable of capturing the semantics of the model could be employed.

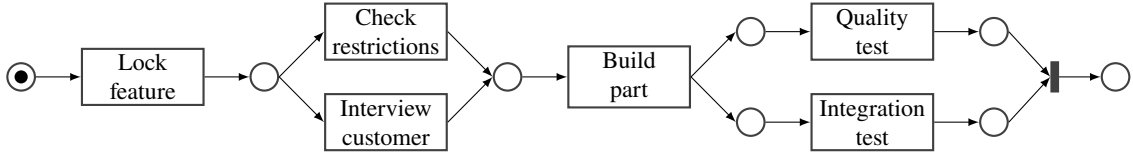

Figure 3: Petri net capturing the behaviour observed in the log from Figure 2.

**Definition 4** (**Labeled Petri net, Workflow net**). A labeled Petri net is a bipartite graph that can be defined by a tuple $N = (P, T, F, \lambda)$, where:

- $P$ is a finite set of places;

- $T$ is a finite set of transitions;

- $P \cap T = \emptyset$;

- $F \subseteq (P \times T) \cup (T \times P)$ is a set of directed arcs; and

- $\lambda : T \to \mathcal{A}$ is a function that assigns to each transition $t \in T$ an activity $a \in \mathcal{A} \cup \emptyset$.

Let $x \in T \cup P$ be a node from the Petri net. We denote $^\bullet x = \{y \mid (y, x) \in F\}$ as the set of inputs of $x$, and $x^\bullet = \{y \mid (x, y) \in F\}$ as the set of outputs of $x$. A labeled Petri net is a *workflow net* (van der Aalst 1998) if and only if:

- $\exists! in \in P : {}^\bullet in = \emptyset$, i.e., there exists a single place in the net with no inputs;

- $\exists! out \in P : out^\bullet = \emptyset$, i.e., there exists a single place in the net with no outputs;

- Adding to the net a transition $t^*$ such that $^\bullet t^* = out$ and $t^{*\bullet} = in$ results in a strongly connected graph, i.e., all $t \in T$ are in a path from $in$ to $out$;

The state of a Petri net, namely a *marking*, is a function $m : P \to \mathbb{N}$ that indicates the number of tokens contained in the place $p \in P$. In a workfow net, the initial marking, denoted as $m_0$, is the one in which only $in$ contains a token. Let $t \in T$ be a transition from the Petri net. We say $t$ is *enabled* when $\forall p \in {}^\bullet t \to m(p) > 0$, i.e., when all its input places have at least one token. When an enabled transition is fired, it consumes a token from each input place and produces a new token in each output place.

We call $B_N$ to the set of all possible paths —i.e., all the unique possible sequences of transition labels— from $in$ to $out$ in the workflow net $N$. For the rest of this paper, when referring to Petri nets, we assume we are talking about workflow nets. Figure 3 shows an example Petri net, where circles represent places, rectangles represent transitions and a black dot represents a token. In this example, the behavior of the model is:

$$B_N = \begin{cases} \langle \text{Lock feature, Check restrictions, Build part, Quality test, Integration test} \rangle \\ \langle \text{Lock feature, Check restrictions, Build part, Integration test, Quality test} \rangle \\ \langle \text{Lock feature, Interview customer, Build part, Quality test, Integration test} \rangle \\ \langle \text{Lock feature, Interview customer, Build part, Integration test, Quality test} \rangle \end{cases}$$

The quality of a process models can be evaluated with respect to a log using *conformance checking metrics*. In this article, we will use two of these metrics as the basis of the algorithm for detecting gradual drifts: *fitness* and *precision*.

**Definition 5** (**Fitness metric**). Fitness measures the fraction of behaviour observed in the log that is captured by the model (Carmona et al. 2018).

$$\gamma(L, N) = \frac{|B_L \cap B_N|}{|B_L|}$$

In the *state-of-the-art* have been proposed a multitude of techniques for fitness computation (Munoz-Gama 2016), that vary in determining the degree of compliance of the model with respect to the log traces (van der Aalst et al. 2012; Adriansyah 2014; de Leoni and van der Aalst 2013). For example, a model with a behavior that only differs in a single activity with respect to a trace with 8 activities can have a fitness of 0 if the metric only considers perfect matches or it can have a fitness of 7/8 if the metric accounts for the activities that do not deviate from the behaviour supported by the model. In this paper, we have opted for a metric that computes the percentage of log traces that are fully supported by the model.

$$\gamma(L, N) = \frac{|\{\tau : \tau \in L \land B_\tau \in B_N\}|}{|L|}$$

For example, considering the log and the model from Figure 2 and Figure 3, respectively:

$$\gamma(L, N) = \frac{|L|}{|L|} = 1$$

because all traces from the log are supported by the model. This metric is quite restrictive when calculating the fitness, but for concept drift detection we just need an estimator of how good the model captures the behaviour observed in the log. As a counterpart, by using this metric we obtain a high performance, which is important since it will be calculated repeatedly as the log is processed.

**Definition 6** (**Precision metric**). Precision measures the fraction of allowed behaviour that is observed in the log (Carmona et al. 2018).

$$\rho(L, N) = \frac{|B_L \cap B_N|}{|B_N|}$$

In the case of precision, many implementations have been also proposed in the *state-of-the-art* (van der Aalst et al. 2012; de Leoni and van der Aalst 2013; vanden Broucke et al. 2014; Rozinat and van der Aalst 2008). Although we could have used any of those implementations, in *CRIER* we consider a custom metric as an estimator of the precision evolution since our algorithm only needs to know whether the precision has changed or not —i.e., it does not need to know the exact value of the precision—. Specifically, this metric checks the percentage of direct relations between activities from the model observed in the traces.

$$\rho(L, N) = \frac{\left|\xrightarrow{L} \cap \xrightarrow{N}\right|}{\left|\xrightarrow{N}\right|}$$

where $\xrightarrow{L}$ are the pairs of consecutive activities found in the traces, and $\xrightarrow{N}$ are the directly connected pairs of activities in the model. For example, using the log and the model from Figure 2 and Figure 3,

respectively:

$$\xrightarrow{L} = \begin{cases} (Lock\ feature \rightarrow Chech\ restrictions), (Lock\ feature \rightarrow Interview\ customer), \\ (Chech\ restrictions \rightarrow Build\ part), (Interview\ customer \rightarrow Build\ part), \\ (Build\ part \rightarrow Integration\ test), (Build\ part \rightarrow Quality\ test), \\ (Integration\ test \rightarrow Quality\ test), (Quality\ test \rightarrow Integration\ test) \end{cases}$$

$$\xrightarrow{N} = \begin{cases} (Lock\ feature \rightarrow Chech\ restrictions), (Lock\ feature \rightarrow Interview\ customer), \\ (Chech\ restrictions \rightarrow Build\ part), (Interview\ customer \rightarrow Build\ part), \\ (Build\ part \rightarrow Integration\ test), (Build\ part \rightarrow Quality\ test) \end{cases}$$

and, therefore, $\rho(L, N) = 6/6 = 1$. But if the relation (*Lock feature* $\rightarrow$ *Chech restrictions*) is not observed in the log, $\rho(L, N) = 5/6 = 0.83$, showing that the estimator decreases when the precision also decreases.

To capture the evolution of the process over time we use a sliding window, which is a sublog containing only the most recent traces at a given instant, being one of the main methods used as a basis for concept drift detection algorithms (Bifet and Gavaldà 2007).

**Definition 7** (**Sliding window**). The sliding window of size $n$ over a log $L$ is defined as $\omega_i(L, n) = \langle \tau_{i-n}, \ldots, \tau_i \rangle$, where its content is conformed by the most recent $n$ traces present in the log at instant $i$. When a new trace is read from the log, the window slides one position, so the oldest trace is forgotten and the new trace is incorporated. This way, the algorithm only takes into account the most recent information for detecting the changes.

The main objective of this paper is the identification of changes in the process structure over time and its classification in sudden and gradual drifts. One of the greatest challenges is differentiating between a real change and an anomalous execution. To accomplish this objective and address this challenge, we use the concept of drift candidate.

**Definition 8** (**Drift candidate**). A drift candidate can be defined as a potential change in the structure of the process that has to be confirmed later. Given two consecutive windows $\omega_i$ and $\omega_{i+1}$, and two process models $N_i = (P_{N_i}, T_{N_i}, F_{N_i}, \lambda_{N_i})$ and $N_{i+1} = (P_{N_{i+1}}, T_{N_{i+1}}, F_{N_{i+1}}, \lambda_{N_{i+1}})$ discovered from each of these windows using the same discovery algorithm, we say that window $\omega_i$ is a drift candidate when $T_{N_i} \neq T_{N_{i+1}} \vee F_{N_i} \neq F_{N_{i+1}}$. A drift is confirmed only after several successive windows are marked as drift candidates and, in that moment, the change is pinpointed to the specific trace that triggered the change —the first trace of the first candidate in the case of fitness, or the last trace of the first window in the case of precision—.

Although there are several types of concept drifts, in this paper we focus on the detection of gradual drifts and on the distinction with sudden drifts.

**Definition 9** (**Gradual drift, Sudden drift**). A gradual drift is defined as a change where the behaviour before the change does not disappear suddenly, but coexists with the one after the change for a period of time, while vanishing until it is no longer observed. Given two time instants $t_1$ and $t_2$ that bound the change, two models can be discovered, $N_{<t_1}$ and $N_{>t_2}$, that capture the behaviour before and after those instants. For a change to be considered as gradual, four conditions must be satisfied:

1. Behaviour captured by the models discovered before $t_1$ and after $t_2$ are different: $B_{N_{<t_1}} \neq B_{N_{>t_2}}$.

2. Some of the behaviour observed between $t_1$ and $t_2$ is captured by the model discovered before $t_1$: $B_{L_{[t_1,t_2]}} \cap B_{N_{<t_1}} \neq \emptyset$.

3. Some of the behaviour observed between $t_1$ and $t_2$ is captured by the model discovered after $t_2$: $B_{L_{[t_1,t_2]}} \cap B_{N_{>t_2}} \neq \emptyset$.

4. All of the behaviour observed between $t_1$ and $t_2$ is captured either by the model discovered before $t_1$, by the model discovered after $t_2$ or by both: $B_{L_{[t_1,t_2]}} \subseteq (B_{N_{<t_1}} \cup B_{N_{>t_2}})$.

In any of these conditions are not met, the change is considered a sudden change, where the old behaviour is replaced instantly by a new one.

## 4 Gradual drift detection using conformance checking

A gradual change is defined between two instants $t_1$ and $t_2$, where the process has a behaviour $B_1$ before $t_1$, a different behaviour $B_2$ after $t_2$, while $B_1$ and $B_2$ coexist in $[t_1, t_2]$ (see Def. 9).

**Theorem 1.** *All gradual drifts are characterized by a fitness change in $t_1$ and by a precision change in $t_2$.*

*Proof.* Consider a log $L$, with a gradual change between instants $t_1$ and $t_2$, in which the new behaviour is first observed at $t_1$ and some old behaviour is less and less frequently observed, until it disappears completely at $t_2$. Let us consider three disjoint sets of behaviour:

- $B_c$, which contains the behaviour that appears throughout the whole duration of the log $L$ and which can be empty;

- $B_p$, which contains the previous behaviour of $L$ that disappears between $t_1$ and $t_2$ and which cannot be empty; and

- $B_n$, which contains the new behaviour that starts to appear after $t_1$ and replaces $B_p$ after $t_2$ and which also cannot be empty.

Using these two instants $t_1$ and $t_2$, we can split the log $L$ in three sublogs. Let $L_{<t_1}$ be the log containing all the traces before $t_1$, $L_{[t_1,t_2]}$ be the log containing the traces between $t_1$ and $t_2$, and $L_{>t_2}$ be the log with the remaining traces after $t_2$. Similarly, we can define the corresponding behaviors for these logs as $B_{L_{<t_1}} = B_c \cup B_p$, $B_{L_{[t_1,t_2]}} = B_c \cup B_p \cup B_n$, and $B_{L_{>t_2}} = B_c \cup B_n$, respectively, and the reference models $N_{<t_1}$, $N_{[t_1,t2]}$ and $N_{>t_2}$ that can be discovered from the traces of these logs, respectively. The values for fitness and precision before the change, i.e., $L_{<t_1}$, can be computed as follows:

$$\gamma(L_{<t_1}, N_{<t_1}) = \frac{|B_{L_{<t_1}} \cap B_{N_{<t_1}}|}{|B_{L_{<t_1}}|} = \frac{|(B_c \cup B_p) \cap B_{N_{<t_1}}|}{|B_c \cup B_p|} = \frac{|(B_c \cap B_{N_{<t_1}}) \cup (B_p \cap B_{N_{<t_1}})|}{|B_c \cup B_p|}$$

$$\rho(L_{<t_1}, N_{<t_1}) = \frac{|B_{L_{<t_1}} \cap B_{N_{<t_1}}|}{|B_{N_{<t_1}}|} = \frac{|(B_c \cup B_p) \cap B_{N_{<t_1}}|}{|B_{N_{<t_1}}|} = \frac{|(B_c \cap B_{N_{<t_1}}) \cup (B_p \cap B_{N_{<t_1}})|}{|B_{N_{<t_1}}|}$$

Which can be simplified since $B_c$ and $B_p$ are disjoint sets:

$$\gamma(L_{<t_1}, N_{<t_1}) = \frac{|B_c \cap B_{N_{<t_1}}| + |B_p \cap B_{N_{<t_1}}|}{|B_c| + |B_p|} = \frac{|B_c \cap B_{N_{<t_1}}|}{|B_c| + |B_p|} + \frac{|B_p \cap B_{N_{<t_1}}|}{|B_c| + |B_p|}$$

$$\rho(L_{<t_1}, N_{<t_1}) = \frac{|B_c \cap B_{N_{<t_1}}| + |B_p \cap B_{N_{<t_1}}|}{|B_{N_{<t_1}}|} = \frac{|B_c \cap B_{N_{<t_1}}|}{|B_{N_{<t_1}}|} + \frac{|B_p \cap B_{N_{<t_1}}|}{|B_{N_{<t_1}}|}$$

Similarly, fitness and precision values in the change interval $[t_1, t_2]$ with respect to this same model $N_{<t_1}$ can be computed as follows:

$$\gamma(L_{[t_1,t_2]}, N_{<t_1}) = \frac{|B_{L_{[t_1,t_2]}} \cap B_{N_{<t_1}}|}{|B_{L_{[t_1,t_2]}}|} = \frac{|(B_c \cup B_p \cup B_n) \cap B_{N_{<t_1}}|}{|B_c \cup B_p \cup B_n|}$$

$$= \frac{|B_c \cap B_{N_{<t_1}}|}{|B_c| + |B_p| + |B_n|} + \frac{|B_p \cap B_{N_{<t_1}}|}{|B_c| + |B_p| + |B_n|} + \frac{|B_n \cap B_{N_{<t_1}}|}{|B_c| + |B_p| + |B_n|}$$

$$\rho(L_{[t_1,t_2]}, N_{<t_1}) = \frac{|B_{L_{[t_1,t_2]}} \cap B_{N_{<t_1}}|}{|B_{N_{<t_1}}|} = \frac{|(B_c \cup B_p \cup B_n) \cap B_{N_{<t_1}}|}{|B_{N_{<t_1}}|}$$

$$= \frac{|B_c \cap B_{N_{<t_1}}|}{|B_{N_{<t_1}}|} + \frac{|B_p \cap B_{N_{<t_1}}|}{|B_{N_{<t_1}}|} + \frac{|B_n \cap B_{N_{<t_1}}|}{|B_{N_{<t_1}}|}$$

This equation can be simplified since $B_n$ only starts to appear after $t_1$, i.e., $B_{N_{<t_1}} \cap B_n = \emptyset$:

$$\gamma(L_{[t_1,t_2]}, N_{<t_1}) = \frac{|B_c \cap B_{N_{<t_1}}|}{|B_c| + |B_p| + |B_n|} + \frac{|B_p \cap B_{N_{<t_1}}|}{|B_c| + |B_p| + |B_n|}$$

$$\rho(L_{[t_1,t_2]}, N_{<t_1}) = \frac{|B_c \cap B_{N_{<t_1}}|}{|B_{N_{<t_1}}|} + \frac{|B_p \cap B_{N_{<t_1}}|}{|B_{N_{<t_1}}|}$$

Moreover, since $B_p \neq \emptyset$ and $B_n \neq \emptyset$, $|B_c| + |B_p| + |B_n| > |B_c| + |B_p|$. Thus, $\gamma(L_{<t_1}, N_{<t_1}) > \gamma(L_{[t_1,t_2]}, N_{<t_1})$ but $\rho(L_{<t_1}, N_{<t_1}) = \rho(L_{[t_1,t_2]}, N_{<t_1})$, confirming our hypothesis that fitness decreases at the beginning of a gradual drift, but precision remaining unaltered.

Once the beginning of the drift has been detected, a new model $N_{[t_1,t_2]}$ is discovered using the log $L_{[t_1,t_2]}$. Then, the conformance metrics can be computed using Def. 5 and Def. 6 as follows:

$$\gamma(L_{[t_1,t_2]}, N_{[t_1,t_2]}) = \frac{|B_{L_{[t_1,t_2]}} \cap B_{N_{[t_1,t_2]}}|}{|B_{L_{[t_1,t_2]}}|} = \frac{|B_c \cap B_{N_{[t_1,t_2]}}|}{|B_c| + |B_p| + |B_n|} + \frac{|B_p \cap B_{N_{[t_1,t_2]}}|}{|B_c| + |B_p| + |B_n|} + \frac{|B_n \cap B_{N_{[t_1,t_2]}}|}{|B_c| + |B_p| + |B_n|}$$

$$\rho(L_{[t_1,t_2]}, N_{[t_1,t_2]}) = \frac{|B_{L_{[t_1,t_2]}} \cap B_{N_{[t_1,t_2]}}|}{|B_{N_{[t_1,t_2]}}|} = \frac{|B_c \cap B_{N_{[t_1,t_2]}}|}{|B_{N_{[t_1,t_2]}}|} + \frac{|B_p \cap B_{N_{[t_1,t_2]}}|}{|B_{N_{[t_1,t_2]}}|} + \frac{|B_n \cap B_{N_{[t_1,t_2]}}|}{|B_{N_{[t_1,t_2]}}|}$$

We can then proceed to define the values after $t_2$, where $B_p$ has totally disappeared:

$$\gamma(L_{>t_2}, N_{[t_1,t_2]}) = \frac{|B_{L_{>t_2}} \cap B_{N_{[t_1,t_2]}}|}{|B_{L_{>t_2}}|} = \frac{|B_c \cap B_{N_{[t_1,t_2]}}|}{|B_c| + |B_n|} + \frac{|B_n \cap B_{N_{[t_1,t_2]}}|}{|B_c| + |B_n|}$$

$$\rho(L_{>t_2}, N_{[t_1,t_2]}) = \frac{|B_{L_{>t_2}} \cap B_{N_{[t_1,t_2]}}|}{|B_{N_{[t_1,t_2]}}|} = \frac{|(B_c \cap B_{N_{[t_1,t_2]}})|}{|B_{N_{[t_1,t_2]}}|} + \frac{|B_n \cap B_{N_{[t_1,t_2]}}|}{|B_{N_{[t_1,t_2]}}|}$$

Nothing can be affirmed with respect to fitness, but we can conclude that $\rho(L_{[t_1,t_2]}, N_{[t_1,t_2]}) > \rho(L_{>t_2}, N_{[t_1,t_2]})$, i.e., precision will decrease at the end of the gradual drift.

A situation in which a gradual change starts with a change in precision at $t_1$ is impossible, since the change in precision necessarily implies a sudden change in the model structure. Since this proof is straightforward, we did not include it in this paper —$B_n = \emptyset$, since no new behavior is added after $t_1$, otherwise a change in fitness should be present, and $B_{L_{<t1}} = B_{L_{[t1,t2]}}$, which contradicts the Def. 8—. □

| ID | Trace | Window | Window fitness | Window precision | Actions | Model |
|---|---|---|---|---|---|---|
| $\tau_1$ | ABCD | $\omega_1 = \langle \tau_1 \rangle$ | – | – | Window is not full. Read new trace | |
| $\tau_2$ | ABCD | $\omega_2 = \langle \tau_1, \tau_2 \rangle$ | – | – | Window is not full. Read new trace | |
| $\tau_3$ | ABCD | $\omega_3 = \langle \tau_1, \tau_2, \tau_3 \rangle$ | – | – | Window is not full. Read new trace | |
| $\tau_4$ | ABCD | $\omega_4 = \langle \tau_1, \tau_2, \tau_3, \tau_4 \rangle$ | $\gamma(N_1, \omega_4) = 1.00$ | $\rho(N_1, \omega_4) = 1.00$ | Discover model for $\omega_4$ | $N_1$ 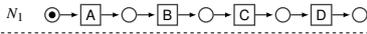 |
| $\tau_5$ | ABCD | $\omega_5 = \langle \tau_2, \tau_3, \tau_4, \tau_5 \rangle$ | $\gamma(N_1, \omega_5) = 1.00$ | $\rho(N_1, \omega_5) = 1.00$ | No drift candidate detected | |
| $\tau_6$ | ABCD | $\omega_6 = \langle \tau_3, \tau_4, \tau_5, \tau_6 \rangle$ | $\gamma(N_1, \omega_6) = 1.00$ | $\rho(N_1, \omega_6) = 1.00$ | No drift candidate detected | |
| $\tau_7$ | ABCD | $\omega_7 = \langle \tau_4, \tau_5, \tau_6, \tau_7 \rangle$ | $\gamma(N_1, \omega_7) = 1.00$ | $\rho(N_1, \omega_7) = 1.00$ | No drift candidate detected | |
| $\tau_8$ | ABCD | $\omega_8 = \langle \tau_5, \tau_6, \tau_7, \tau_8 \rangle$ | $\gamma(N_1, \omega_8) = 1.00$ | $\rho(N_1, \omega_8) = 1.00$ | No drift candidate detected | |
| $\tau_9$ | ABDC | $\omega_9 = \langle \tau_6, \tau_7, \tau_8, \tau_9 \rangle$ | $\boldsymbol{\gamma N_1, \omega_9 = 0.75}$ | $\rho(N_1, \omega_9) = 1.00$ | Drift candidate on $\omega_9$ (by fitness) | |
| $\tau_{10}$ | ABCD | $\omega_{10} = \langle \tau_7, \tau_8, \tau_9, \tau_{10} \rangle$ | $\boldsymbol{\gamma N_1, \omega_{10} = 0.75}$ | $\rho(N_1, \omega_{10}) = 1.00$ | Drift candidate on $\omega_{10}$ (by fitness) | |
| $\tau_{11}$ | ABDC | $\omega_{11} = \langle \tau_8, \tau_9, \tau_{10}, \tau_{11} \rangle$ | $\boldsymbol{\gamma N_1, \omega_{11} = 0.50}$ | $\rho(N_1, \omega_{11}) = 1.00$ | Drift candidate on $\omega_{11}$ (by fitness) | |
| $\tau_{12}$ | ABCD | $\omega_{12} = \langle \tau_9, \tau_{10}, \tau_{11}, \tau_{12} \rangle$ | $\boldsymbol{\gamma N_1, \omega_{12} = 0.50}$ | $\rho(N_1, \omega_{12}) = 1.00$ | Drift candidate on $\omega_{12}$ (by fitness) Confirm drift at $\tau_9$ (by fitness) Discover model for $\omega_{12}$ | $N_2$ 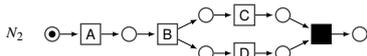 |
| $\tau_{13}$ | ABDC | $\omega_{13} = \langle \tau_{10}, \tau_{11}, \tau_{12}, \tau_{13} \rangle$ | $\gamma(N_2, \omega_{13}) = 1.00$ | $\rho(N_2, \omega_{13}) = 1.00$ | No drift candidate detected | |
| $\tau_{14}$ | ABCD | $\omega_{14} = \langle \tau_{11}, \tau_{12}, \tau_{13}, \tau_{14} \rangle$ | $\gamma(N_2, \omega_{14}) = 1.00$ | $\rho(N_2, \omega_{14}) = 1.00$ | No drift candidate detected | |
| $\tau_{15}$ | ABDC | $\omega_{15} = \langle \tau_{12}, \tau_{13}, \tau_{14}, \tau_{15} \rangle$ | $\gamma(N_2, \omega_{15}) = 1.00$ | $\rho(N_2, \omega_{15}) = 1.00$ | No drift candidate detected | |
| $\tau_{16}$ | ABCD | $\omega_{16} = \langle \tau_{13}, \tau_{14}, \tau_{15}, \tau_{16} \rangle$ | $\gamma(N_2, \omega_{16}) = 1.00$ | $\rho(N_2, \omega_{16}) = 1.00$ | No drift candidate detected | |
| $\tau_{17}$ | ABDC | $\omega_{17} = \langle \tau_{14}, \tau_{15}, \tau_{16}, \tau_{17} \rangle$ | $\gamma(N_2, \omega_{17}) = 1.00$ | $\rho(N_2, \omega_{17}) = 1.00$ | No drift candidate detected | |
| $\tau_{18}$ | ABDC | $\omega_{18} = \langle \tau_{15}, \tau_{16}, \tau_{17}, \tau_{18} \rangle$ | $\gamma(N_2, \omega_{18}) = 1.00$ | $\rho(N_2, \omega_{18}) = 1.00$ | No drift candidate detected | |
| $\tau_{19}$ | ABDC | $\omega_{19} = \langle \tau_{16}, \tau_{17}, \tau_{18}, \tau_{19} \rangle$ | $\gamma(N_2, \omega_{19}) = 1.00$ | $\rho(N_2, \omega_{19}) = 1.00$ | No drift candidate detected | |
| $\tau_{20}$ | ABDC | $\omega_{20} = \langle \tau_{17}, \tau_{18}, \tau_{19}, \tau_{20} \rangle$ | $\gamma(N_2, \omega_{20}) = 1.00$ | $\boldsymbol{\rho N_2, \omega_{20} = 0.66}$ | Drift candidate on $\omega_{20}$ (by precision) | |
| $\tau_{21}$ | ABDC | $\omega_{21} = \langle \tau_{18}, \tau_{19}, \tau_{20}, \tau_{21} \rangle$ | $\gamma(N_2, \omega_{21}) = 1.00$ | $\boldsymbol{\rho N_2, \omega_{21} = 0.66}$ | Drift candidate on $\omega_{21}$ (by precision) | |
| $\tau_{22}$ | ABDC | $\omega_{22} = \langle \tau_{19}, \tau_{20}, \tau_{21}, \tau_{22} \rangle$ | $\gamma(N_2, \omega_{22}) = 1.00$ | $\boldsymbol{\rho N_2, \omega_{22} = 0.66}$ | Drift candidate on $\omega_{22}$ (by precision) | |
| $\tau_{23}$ | ABDC | $\omega_{23} = \langle \tau_{20}, \tau_{21}, \tau_{22}, \tau_{23} \rangle$ | $\gamma(N_2, \omega_{23}) = 1.00$ | $\boldsymbol{\rho N_2, \omega_{23} = 0.66}$ | Drift candidate on $\omega_{23}$ (by precision) Confirm drift at $\tau_{17}$ (by precision) Discover model for $\omega_{23}$ | $N_3$ 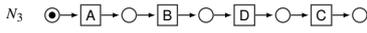 |
| $\tau_{24}$ | ABDC | $\omega_{24} = \langle \tau_{21}, \tau_{22}, \tau_{23}, \tau_{24} \rangle$ | $\gamma(N_3, \omega_{24}) = 1.00$ | $\rho(N_3, \omega_{24}) = 1.00$ | No drift candidate detected | |
| $\tau_{25}$ | ABDC | $\omega_{25} = \langle \tau_{22}, \tau_{23}, \tau_{24}, \tau_{25} \rangle$ | $\gamma(N_3, \omega_{25}) = 1.00$ | $\rho(N_3, \omega_{25}) = 1.00$ | No drift candidate detected | |
| $\tau_{26}$ | ABDC | $\omega_{26} = \langle \tau_{23}, \tau_{24}, \tau_{25}, \tau_{26} \rangle$ | $\gamma(N_3, \omega_{26}) = 1.00$ | $\rho(N_3, \omega_{26}) = 1.00$ | No drift candidate detected | |
| $\tau_{27}$ | ABDC | $\omega_{27} = \langle \tau_{24}, \tau_{25}, \tau_{26}, \tau_{27} \rangle$ | $\gamma(N_3, \omega_{27}) = 1.00$ | $\rho(N_3, \omega_{27}) = 1.00$ | No drift candidate detected | |

Figure 4: Example of a gradual change where some previously unobserved behavior starts to replace the previous one at $\tau_9$, which is no longer observed after trace $\tau_{16}$.

## 5 Algorithm for gradual drift detection

The premise underlying the operation of *CRIER* is that, using fitness and precision metrics, gradual drifts can be detected with high accuracy. However, to better understand how the algorithm works, we illustrate the gradual drift detection with the example of Figure 4. The example is based on a sliding window of size 4. First, traces are read until the window is full. Then, traces from the first full window $\omega_4$ are mined with a discovery algorithm to obtain the model that describes its behavior. Then, the window is shifted trace by trace and for each shift the fitness and precision —which are initially 1— are checked to detect whether they change. In the window $\omega_9$ the fitness decreases while the precision does not change, so $\tau_9$ is labeled as a drift candidate. The fitness decrease lasts for three windows, which means that traces from $\tau_{10}$ to $\tau_{12}$ are also drift candidates, so trace $\tau_9$ is finally labeled as a drift. As a consequence, a new model is discovered at $\omega_{12}$, starting a new cycle of the algorithm, until a change in fitness or precision is detected in the next windows. In this example, a precision decrease is detected for the windows $\omega_{20}$, $\omega_{21}$, $\omega_{22}$ and $\omega_{23}$, confirming a drift in trace $\tau_{17}$ and, therefore, a gradual drift between traces $\tau_9$ and $\tau_{17}$.

**Algorithm 1** CRIER (Conformance-based GRadual DrIft DetEction AlgoRithm)

    **Inputs:** an event log $L$, minimum size of the sliding window $n'$
    **Outputs:** a set of traces (for sudden changes)/segments of the log (for gradual changes) causing drifts

1: **procedure** CONCEPTDRIFTDETECTION($L, n'$)
2:    $D^S, D^G \leftarrow [\ ]$ //confirmed sudden and gradual drifts
3:    $n \leftarrow$ ADJUSTWINDOW($n', \langle \tau_1, \ldots, \tau_{|L|} \rangle$)
4:    $i \leftarrow n$
5:    $\omega_i \leftarrow \langle \tau_{i-n}, \ldots, \tau_i \rangle$
6:    $N \leftarrow [discover(\omega_i)]$ //save the model in the model history
7:    $\tau^*, \tau' \leftarrow \tau_1$
8:    **while** $i < |L|$ **do**
9:       $\Gamma, P \leftarrow [\ ]$ //fitness and precision measures (Def. 5 and Def. 6)
10:      $D^\Gamma, D^P \leftarrow [\ ]$ //drift candidates (fitness and precision)
11:      $flag \leftarrow FALSE$
12:      **while** $(i < |L|) \land \neg flag$ **do**
13:         $\Gamma \leftarrow \Gamma :: \gamma(\omega_i, N_{|N|})$ //append current fitness
14:         $P \leftarrow P :: \rho(\omega_i, N_{|N|})$ //append current precision
15:         $D^\Gamma \leftarrow D^\Gamma ::$ IDENTIFYDRIFTCANDIDATE($n, \Gamma, D^\Gamma$)
16:         $D^P \leftarrow D^P ::$ IDENTIFYDRIFTCANDIDATE($n, P, D^P$)
17:         **if** CONFIRMDRIFT($n, D^\Gamma$) $\lor$ CONFIRMDRIFT($n, D^P$) **then**//change confirmed
18:            **if** CONFIRMDRIFT($n, D^\Gamma$) **then**
19:               $\tau^* \leftarrow \tau_{i-n}$ //confirmed drift in the last trace from the first candidate
20:            **else if** CONFIRMDRIFT($n, D^P$) **then**
21:               $\tau^* \leftarrow \tau_{i-2n}$ //confirmed drift in the first trace from the first candidate
22:            **end if**
23:            $flag \leftarrow TRUE$
24:            $L' \leftarrow \langle \tau', \ldots, \tau^* \rangle$ //store the sublog between confirmed drifts
25:            $n \leftarrow$ ADJUSTWINDOW($n', \langle \tau_{i+1}, \ldots, \tau_{|L|} \rangle$)
26:            $i \leftarrow i + n$
27:            $\omega_i \leftarrow \langle \tau_{i-n}, \ldots, \tau_i \rangle$
28:            $N \leftarrow N :: discover(\omega_i)$ //append current model to model history
29:            **if** $|N| > 2 \land (\exists \tau \in L' : B_\tau \in B_{N_{|N|-2}}) \land (\exists \tau \in L' : B_\tau \in B_{N_{|N|}}) \land (\forall \tau \in L' : B_\tau \in B_{N_{|N|-2}} \cup B_{N_{|N|}})$ **then**
30:               $\tau' \leftarrow D^S_{|D^S|}$
31:               $D^S \leftarrow \{d \in D^S : d \neq \tau'\}$
32:               $D^G \leftarrow D^G :: [\tau', \tau^*]$ //a gradual change
33:            **else**
34:               $D^S \leftarrow D^S :: \tau^*$ //a sudden change
35:            **end if**
36:            $\tau' \leftarrow \tau^*$
37:         **else**
38:            $i \leftarrow i + 1$
39:            $\omega_i \leftarrow \langle \tau_{i-n}, \ldots, \tau_i \rangle$
40:         **end if**
41:      **end while**
42:    **end while**
43:    **return** $D^S \cup D^G$
44: **end procedure**

## 5.1 Algorithm description

Algorithm 1 shows the pseudocode of the *CRIER* algorithm. The only mandatory inputs are the event log and a minimum window size. As part of the initialization phase, the algorithm creates two empty lists for the confirmed drifts, one for sudden and one for gradual drifts (line 2) and sets the initial window index $i = 1$ (line 4). As the algorithm is based on a sliding window $\omega_i$ (Def. 7), in this initialization

**Algorithm 2** Auxiliary functions

```
 1: function ADJUSTWINDOW(n, L)
 2:     N_1, N_2, N_3 ← ∅
 3:     n' ← n
 4:     while 3n < |L| do
 5:         N_1 ← discover(⟨τ_0, ..., τ_{n'}⟩)
 6:         N_2 ← discover(⟨τ_{n'}, ..., τ_{2n'}⟩)
 7:         N_3 ← discover(⟨τ_{2n'}, ..., τ_{3n'}⟩)
 8:         if B_{N_1} = B_{N_2} = B_{N_3} then
 9:             n' ← increment(n')
10:         else
11:             return n'
12:         end if
13:     end while
14:     return n'
15: end function

16: function IDENTIFYDRIFTCANDIDATE(n, data, D)
17:     Υ ← regress({data_{|data|−n}, ..., data_{|data|}})
18:     m^< ← Υ.slope < 0 ∧ Υ.confidence < 0.05
19:     m^> ← Υ.slope > 0 ∧ Υ.confidence < 0.05
20:     m^= ← (¬m^<) ∧ (¬m^>)
21:     return (|data| > n) ∧ (m^< ∨ m^> ∨ (m^= ∧ (D_{|D|} = true)))
22: end function

23: function CONFIRMDRIFT(n, D)
24:     d' ← ∀d ∈ {D_{|D|−n}, ..., D_{|D|}} : d = true
25:     return (|D| ≥ n) ∧ d'
26: end function
```

phase its optimal size $n$ is computed (line 3) using the function ADJUSTWINDOW (Algorithm 2, lines 1 to 15). The adjustment of the window size is based on the comparison of the behaviour observed in three consecutive windows. In this step, we start with a size $n'$ initialized as the minimum window size $n$ and discover three models from three consecutive windows of size $n'$ (Algorithm 2, lines 5 to 7). If the three models capture the same behaviour (Algorithm 2, line 8), the size of each window $n'$ is incremented by $n'$ (Algorithm 2, line 9) and the process starts again. The procedure finishes when one of the three models has a different behavior or when the condition $3 * n' < |L|$ fails, meaning that it is not possible to discover three consecutive models, returning the last $n'$ as the adjusted optimal window size (Algorithm 2, line 14).

Once the size is adjusted, the window is populated with the first $n$ remaining traces from the initial index. Then, the first process model is discovered (line 6) using the content from this $\omega_i$ window. This model is stored in a list $N$ which contains the set of models that will be discovered for each detected drift.

After this initialization, the drift detection step is executed (lines 8 to 42). This step is composed of a loop that repeats until no traces remain unprocessed in the log. Thus, as part of the initialization performed after each change detection, two empty lists are created (line 9), $\Gamma$ and $P$, for storing the fitness and precision, respectively. Also, two more lists, $D^\Gamma$ and $D^P$, are initialized (line 10). These lists will later be populated with booleans indicating if the successive windows are marked as drift candidates. Finally, a flag used for indicating if a change has been detected is initialized (line 11).

Once this initialization is completed, the main detection loop begins (lines 12 to 41), where the detection is performed based on the trends in the values of the conformance metrics. This loop processes the remaining windows, starting at index *i*, until a new change is detected or to the end of the log if no change is present (line 12). For each window $\omega_i$ the fitness and the precision of the last discovered model $N_{|N|}$ are computed, and their values are appended to $\Gamma$ and $P$, respectively (lines 13 and 14). Then, using the function IDENTIFYDRIFTCANDIDATE from Algorithm 2, the window $\omega_i$ is classified or not as a drift candidate (lines 15 to 16). To carry out this classification step, a simple linear regression —using the ordinary least squares method— is computed over the metric values (Algorithm 2, line 17). If there are enough data for computing the regression and the slope of the fitted line is different from 0 with enough statistical confidence —$p$-value $< 0.05$ in the $t$-test, where $H_0$ states that the slope of the regression is 0—, or the slope is 0 and the previous window has been marked as a drift candidate, the window $\omega_i$ will be marked as a drift candidate. To finalize this detection step the function CONFIRMDRIFT from Algorithm 2 is executed (line 29). This function is in charge of checking whether a drift candidate persists over time, becoming a real change, or, on the contrary, only represents a noisy trace. A drift candidate is confirmed as a real change if the last *n* windows have been also marked as drift candidates (Algorithm 2, lines 23 to 26).

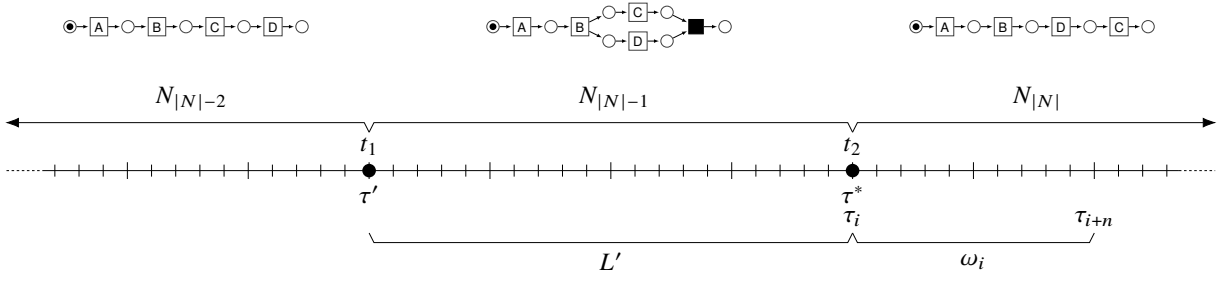

Figure 5: Drift classification phase in *CRIER*.

Once the change is confirmed, it should be pinpointed in time to an specific trace. In case the drift comes from a change in fitness, the trace causing the drift is the last trace from the first window identified as a candidate —i.e., the first trace causing a change in the metric—. In the case of a change in precision, the trace causing the drift is the first trace from the first window identified as candidate —i.e., the first trace from the first window causing a change in precision—. Then, the flag indicating a detection is updated (line 23) and the sublog $L'$ between confirmed drifts is stored for later. The drift must then be classified as sudden or gradual. Figure 5 illustrates this part of the algorithm. The first step is to update both the index at which the next window will start and its optimal size (lines 25 and 26). Then, the new window is populated with the traces and the model describing the behavior contained in this window is discovered (lines 27 and 28). This model is stored in the model list. Since gradual changes are delimited by two drifts, the last three models of this list —$N_{|N|-2}$, $N_{|N|-1}$, and $N_{|N|}$— are used to check the conditions that determine whether the drift is gradual or not. Note that the model $N_{|N|-1}$ corresponds with the sublog in which the confirmed drift has been detected. These conditions are the following (line 29):

1. There are more than two models in the model list ($|N| > 2$).

2. At least, one trace from $L'$ is supported by $N_{|N|}$.

3. At least, one trace from $L'$ is supported by $N_{|N|-2}$.

4. The behaviour from every trace in $L'$ is part of the behaviour of $N_{|N|-2}$ or $N_{|N|}$.

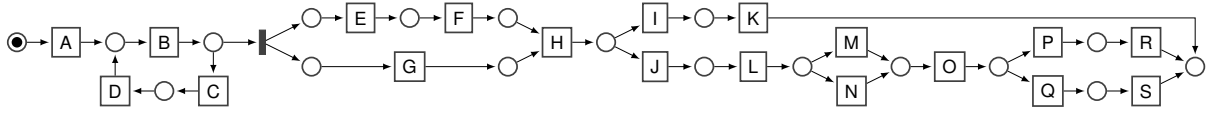

Figure 6: Petri net model used in the validation of the algorithm, where activity names are shortened for better understanbility.

Table 1: Change patterns applied to the original model from Figure 6 and resulting models.

(a) Change patterns.

| Code | Change pattern | Class |
|---|---|---|
| cm | Move fragment into/out of conditional branch | I |
| cp | Duplicate fragment | I |
| pm | Move fragment into/out of parallel branch | I |
| re | Add/remove fragment | I |
| rp | Substitute fragment | I |
| sw | Swap two fragments | I |
| cb | Make fragment skippable/non-skippable | O |
| lp | Make fragments loopable/non-loopable | O |
| cd | Synchronize two fragments | R |
| cf | Make two fragments conditional/sequential | R |
| pl | Make two fragments parallel/sequential | R |

(b) Derived models.

| Model code | Applied change patterns |
|---|---|
| cp | cp |
| pm | pm |
| re | re |
| rp | rp |
| sw | sw |
| cf | cf |
| OIR | lp + re + cd |
| ORI | lp + pl + re |
| RIO | cf + cp + cb |
| ROI | pl + lp + rp |

If $L'$ fulfills these four requirements, the change is classified as gradual, indicating that the drift starts at the trace where the last sudden drift $D^S{}_{|D^S|}$ was detected and lasts until $\tau^*$. Furthermore, $D^S{}_{|D^S|}$ is removed from the sudden drift list (line 31), and appended to the list $D^\Gamma$ of classified drifts (line 32). On the other hand, if the candidate cannot be classified as gradual because it does not meet any of the requirements, the change at $\tau^*$ is classified as sudden (line 34). Finally, the index $i$ is incremented by 1 (line 38), the sliding window is updated to this index (line 39), and the cycle starts again.

# 6 Experimentation

## 6.1 Validation Data

Concept drift algorithms are validated with synthetic data generated from real processes, in which changes are introduced at a specific moment and with a specific duration, since there are no real logs in which these change regions are identified. In this paper, *CRIER* has been validated with synthetic logs generated from a loan granting process, which is the *de-facto* benchmark used to validate concept drift approaches in process mining (Seeliger et al. 2017; Zheng et al. 2017; Maaradji et al. 2017; Yeshchenko et al. 2019). Its model, represented in Figure 6, consist of 19 different activities structured according to typical control constructs such as sequences, parallels, and choices (Dumas et al. 2018).

From this base model, and applying the patterns of change presented by Weber et al. (2008), modified models have been generated using the methodology described by Maaradji et al. (2017). These patterns can be classified into three categories (Table 1a): *(i)* changes involving the insertion of new behavior —marked with *I*—; *(ii)* changes involving the optionalization of a part of the model —marked with *O*—; and *(iii)* changes involving the restructuring and rearrangement of some parts of the model —marked with *R*—. In addition, logs combining the different patterns have also been generated. Only the patterns or patterns combination that can generate a gradual change —i.e., those that add some new, unobserved behaviour, forcing a change in fitness— have been applied to create the validation models, as they are the only ones that can generate a gradual drift (Table 1b). Note that the other patterns —specifically, cm, cb, lp, cd, and pl— produce models that only would change the precision and, therefore, no gradual changes could be generated from these models (see Theorem 1). As a result of this procedure, 10 derived models are created. These models are used to generate the trace logs with which the approach is validated since each one is the result of applying the gradual change that takes place from the base model. For this generation, a cumulative probability function (*cdf*) selects the model —base or derived— that will generate each trace according to a given probability distribution (*P*). Specifically, the following procedure is applied to create the validation logs:

1. The base model $M_1$ —i.e., the loan application process model—, the modified model $M_2$ —i.e., one of the 10 derived models—, and the probability distribution *P* are choosen.

2. A block of 500 traces corresponding to model $M_1$ is generated.

3. While the *cdf* of *P* is below the stopping criterion —$cdf(P) \leq 0.999$—, a model is choosen between $M_1$ and $M_2$ with probabilities $1 - cdf(P)$ and $cdf(P)$, respectively, and a new trace is generated from the selected model. Note that this stopping criterion is due to the fact that the cumulative probability function for some distributions is asymptotic to 1.

4. Repeat from step 2 interchanging $M_1$ and $M_2$.

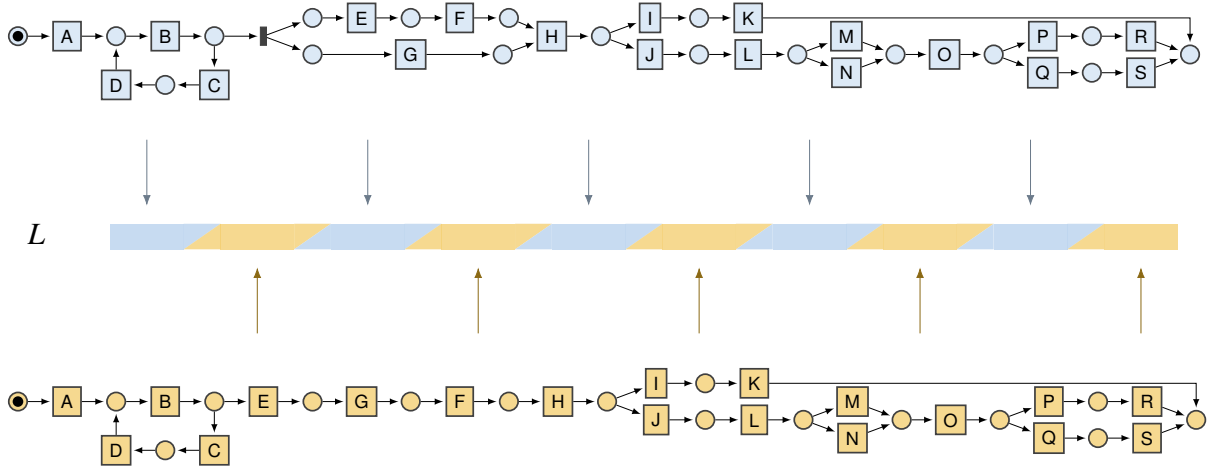

Figure 7: Log generation example.

Figure 7 shows an example of log generation where the base model is combined with a model derived from applying the pm pattern by using a linear probability distribution.

In our experimentation, 4 different probability functions have been considered: (i) *linear*, where the frecuency of the new behaviour increment linearly when new traces are observed; (ii) *Gaussian*, where the change starts to appear slowly and accelerates to end with the old behavior fading away slowly as well; (iii) *exponential*, where the frequency of the new behavior grows very fast at the beginning and slows down as new traces are observed; and (iv) *constant*, where the old and the new behavior have a constant probability of appearing that is not modified during a number of traces. For these distributions, several cumulative probability functions have been applied with different configuration parameters, for a total of 12 distinct probability distributions, as Figure 8 shows. In summary, 120 synthetic logs[1] have been generated, using 10 different patterns and 12 probability distributions. Table 2 summarizes the features of the validation logs, with the log size, the change points and the distributions applied.

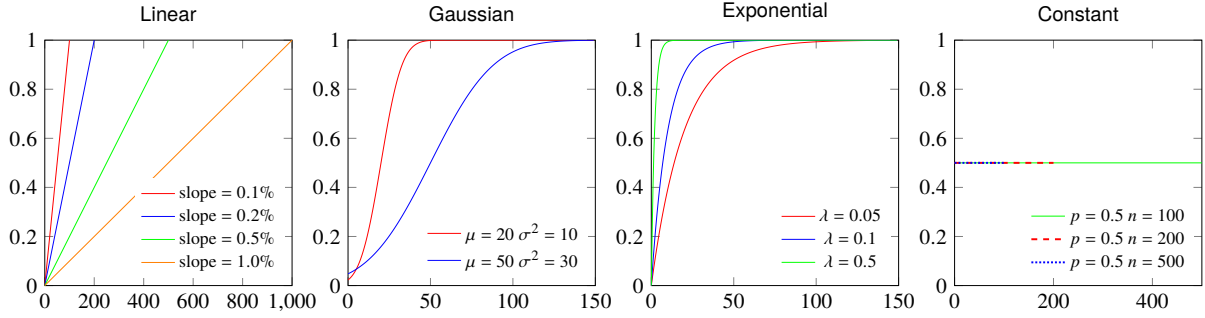

Figure 8: Cumulative distribution functions for the applied distributions in the log generation.

Table 2: Summary of validation logs and their change points, where $t_1$ and $t_2$ are the temporal points between which the gradual change takes place.

| Distribution | Log size | | Change regions | | | | | | | | |
|---|---|---|---|---|---|---|---|---|---|---|---|
| | | | Drift 1 | Drift 2 | Drift 3 | Drift 4 | Drift 5 | Drift 6 | Drift 7 | Drift 8 | Drift 9 |
| Linear (slope = 0.1%) | 14000 | $t_1$ | 500 | 2,000 | 3,500 | 5,000 | 6,500 | 8,000 | 9,500 | 11,000 | 12,500 |
| | | $t_2$ | 1,500 | 3,000 | 4,500 | 6,000 | 7,500 | 9,000 | 10,500 | 12,000 | 13,500 |
| Linear (slope = 0.2%) | 9500 | $t_1$ | 500 | 1,500 | 2,500 | 3,500 | 4,500 | 5,500 | 6,500 | 7,500 | 8,500 |
| | | $t_2$ | 1,000 | 2,000 | 3,000 | 4,000 | 5,000 | 6,000 | 7,000 | 8,000 | 9,000 |
| Linear (slope = 0.5%) | 6800 | $t_1$ | 500 | 1,200 | 1,900 | 2,600 | 3,300 | 4,000 | 4,700 | 5,400 | 6,100 |
| | | $t_2$ | 700 | 1,400 | 2,100 | 2,800 | 3,500 | 4,200 | 4,900 | 5,600 | 6,300 |
| Linear (slope = 1.0%) | 5900 | $t_1$ | 500 | 1,100 | 1,700 | 2,300 | 2,900 | 3,500 | 4,100 | 4,700 | 5,300 |
| | | $t_2$ | 600 | 1,200 | 1,800 | 2,400 | 3,000 | 3,600 | 4,200 | 4,800 | 5,400 |
| Gaussian ($\mu = 20, \sigma^2 = 10$) | 5459 | $t_1$ | 500 | 1,051 | 1,602 | 2,153 | 2,704 | 3,255 | 3,806 | 4,357 | 4,908 |
| | | $t_2$ | 551 | 1,102 | 1,653 | 2,204 | 2,755 | 3,306 | 3,857 | 4,408 | 9,959 |
| Gaussian ($\mu = 50, \sigma^2 = 30$) | 6287 | $t_1$ | 500 | 1,143 | 1,786 | 2,429 | 3,072 | 3,715 | 4,358 | 5,001 | 5,644 |
| | | $t_2$ | 643 | 1,286 | 1,929 | 2,572 | 3,215 | 3,858 | 4,501 | 5,144 | 5,787 |
| Exponential ($\lambda = 0.05$) | 6251 | $t_1$ | 500 | 1,139 | 1,778 | 2,417 | 3,056 | 3,695 | 4,334 | 4,973 | 5,612 |
| | | $t_2$ | 639 | 1,278 | 1,917 | 2,556 | 3,195 | 3,834 | 4,473 | 5,112 | 5,751 |
| Exponential ($\lambda = 0.1$) | 5630 | $t_1$ | 500 | 1,070 | 1,640 | 2,210 | 2,780 | 3,350 | 3,920 | 4,490 | 5,060 |
| | | $t_2$ | 570 | 1,140 | 1,710 | 2,280 | 2,850 | 3,420 | 3,990 | 4,560 | 5,130 |
| Exponential ($\lambda = 0.5$) | 5126 | $t_1$ | 500 | 1,014 | 1,528 | 2,042 | 2,556 | 3,070 | 3,584 | 4,098 | 4,612 |
| | | $t_2$ | 514 | 1,028 | 1,542 | 2,056 | 2,570 | 3,084 | 3,598 | 4,112 | 4,626 |
| Constant ($p = 0.5, n = 100$) | 5900 | $t_1$ | 500 | 1,100 | 1,700 | 2,300 | 2,900 | 3,500 | 4,100 | 4,700 | 5,300 |
| | | $t_2$ | 600 | 1,200 | 1,800 | 2,400 | 3,000 | 3,600 | 4,200 | 4,800 | 5,400 |
| Constant ($p = 0.5, n = 200$) | 6800 | $t_1$ | 500 | 1,200 | 1,900 | 2,600 | 3,300 | 4,000 | 4,700 | 5,400 | 6,100 |
| | | $t_2$ | 700 | 1,400 | 2,100 | 2,800 | 3,500 | 4,200 | 4,900 | 5,600 | 6,300 |
| Constant ($p = 0.5, n = 500$) | 9500 | $t_1$ | 500 | 1,500 | 2,500 | 3,500 | 4,500 | 5,500 | 6,500 | 7,500 | 8,500 |
| | | $t_2$ | 1,000 | 2,000 | 3,000 | 4,000 | 5,000 | 6,000 | 7,000 | 8,000 | 9,000 |

## 6.2 Metrics

When evaluating the performance of a drift detection solution, there are two main metrics used in the *state-of-the-art*. The first metric, called $F_{score}$, calculates the harmonic mean between *precision* and

---

[1]The generated syntetic logs are available at https://gitlab.citius.usc.es/ProcessMining/logs/-/tree/master/drift/gradual

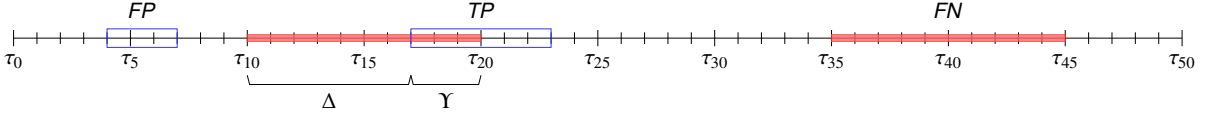

Figure 9: Example of classification of results over a log with two gradual changes. Real drift regions are marked in red. Detected drift regions are outlined in blue.

*recall*, allowing to evaluate how reliable the results are, based on the number of true/false positives and negatives. For gradual changes, a *true positive* ($TP$) is any change whose detection area overlaps with a real change not previously detected, while a *false positive* ($FP$) is any change that does not correspond to a region of real change, or that matches a region of change previously detected. In addition, a *false negative* ($FN$) is a region of change that does not overlap with anyone of the detected changes. Considering this, the $F_{score}$ metric is defined as follows:

$$F_{score} = \frac{2 \times precision \times recall}{precision + recall}$$

$$precision = \frac{TP}{TP + FP}$$

$$recall = \frac{TP}{TP + FN}$$

The second metric, called *delay* ($\Delta$), measures how late a drift is reported from the actual occurrence until its detection. In the case of gradual changes, since we are dealing with areas of change and not single points, we measure the delay in detecting the beginning of the region.

$$\Delta(d^R, d^D) = |\min(d^R) - \min(d^D)|$$

where $d^R$ and $d^D$ are intervals indicating a real and a detected drift region, respectively.

Finally, and to better reflect the goodness of the results when dealing with gradual changes, these two metrics are complemented with a third one: the *change region overlapping* ($\Upsilon$). This metric evaluates the percentage of the actual region of change that has been detected by the algorithm, reflecting how well the duration of the changes is captured.

$$\Upsilon(d^R, d^D) = \frac{|d^R \cap d^D|}{|d^R|}$$

where $d^R$ and $d^D$ are intervals indicating a real and a detected drift region, respectively.

Figure 9 shows an example of how these metrics are applied in the evaluation of the results. This example presents a log with two gradual changes, one between traces $\tau_{10}$ and $\tau_{20}$, and the other one between traces $\tau_{35}$ and $\tau_{45}$. The algorithm detects two changes, one between traces $\tau_4$ and $\tau_7$, classified as a false positive as no real change happened in this time interval, and one between traces $\tau_{17}$ and $\tau_{23}$, classified as a true positive because it overlaps a real change. In this second detection, the delay would be 7 traces —from trace $\tau_{10}$ to trace $\tau_{17}$—. On the other hand, the change region overlapping would be 30 %, since only 3 traces out of the 10 that constitute the region of change are detected. Finally, the second change, between traces $\tau_{35}$ and $\tau_{45}$, will be classified as a false negative, since the algorithm has not detected any change in that region.

Table 3: Evidence levels proposed by Jeffreys (1998) for the Bayes factors.

| | **Bayes factor** $BF_{H_i H_j}$ | | | **Interpretation** |
|---|---|---|---|---|
| | $BF_{H_i H_j}$ | > | 100 | Extreme evidence for $H_i$ |
| 100 > | $BF_{H_i H_j}$ | > | 30 | Very strong evidence for $H_i$ |
| 30 > | $BF_{H_i H_j}$ | > | 10 | Strong evidence for $H_i$ |
| 10 > | $BF_{H_i H_j}$ | > | 3 | Moderate evidence for $H_i$ |
| 3 > | $BF_{H_i H_j}$ | > | 1 | Anecdotal evidence for $H_i$ |
| | $BF_{H_i H_j}$ | = | 1 | No evidence |
| 1 > | $BF_{H_i H_j}$ | > | 0.3333 | Anecdotal evidence for $H_j$ |
| 0.3333 > | $BF_{H_i H_j}$ | > | 0.1 | Moderate evidence for $H_j$ |
| 0.1 > | $BF_{H_i H_j}$ | > | 0.0333 | Strong evidence for $H_j$ |
| 0.0333 > | $BF_{H_i H_j}$ | > | 0.01 | Very strong evidence for $H_j$ |
| 0.01 > | $BF_{H_i H_j}$ | | | Extreme evidence for $H_j$ |

Furthermore, results of the different algorithms have been also evaluated using a *Bayesian hypothesis test*. We opted for a Bayesian approach instead of a traditional null hypothesis significance test (*NHST*) (Ly et al. 2020) because:

1. *NHST* does not provide any certainty about the validity of the null hypothesis. Thus, if the hypothesis is not rejected, it can only be stablished that there is no evidence to reject it, but never to accept it as such.

2. *NHST* does not estimate the probability of the hypotheses to be valid, so comparing the algorithms is harder.

In this paper, we have performed a *BAyesian INformative Hypothesis Evaluation* (*BAIN*) (Gu et al. 2018) to analyze the experimentation. *BAIN* is based on the use of *Bayes factors* to compare the conditional probability between two competing hypotheses. In a simplified form, a *Bayesian factor* $BF_{H_i H_j}$ is the ratio of the probability of the hypothesis $H_i$ to the probability of the hypothesis $H_j$. For example, $BF_{H_i H_j} = 5$ would indicate that the hypothesis $H_i$ is 5 times more likely than hypothesis $H_j$. The most frequent interpretation of these Bayes factors was proposed by Jeffreys (1998), where the Bayes factor values are classified in 11 levels, from *no evidence* to *extreme evidence*. Table 3 shows the different levels of evidence defined in this clasification. For all the hypothesis proposed for the test, symbols >, = and < refer to the algorithm obtaining better, equal or worst results. All tests have been executed using the software package JASP (Love et al. 2019).

## 6.3 Set up

The results of *CRIER* have been compared with the two main publicly available *state-of-the-art* algorithms: the proposal from *Martjushev et al.* 2015 —available as a plugin of the process mining platform ProM[2]— and *ProDrift*[3] (Maaradji et al. 2017). Note that *ProDrift* has not been run for logs with Gaussian, exponential, and constant distributions since authors explicitly state that it only deals with linear changes. The rest of the approaches identified in the *state-of-the-art* have not been tested as the source code of their

---
[2]https://www.promtools.org/
[3]https://kodu.ut.ee/ dumas/tools/ProDrift2.5.zip

implementations are not available. The specific configurations used for these two algorithms were as follows:

1. *Martjushev et al.*: local features using the follows relations for all pairs of activities; *J-measure* with a window size of 10; adaptive window with a minimum size of 50 and a maximum of 500; a step of 1, and an automatic gap size; and comparing windows with the *Kolmogorov-Smirnov* test with a p-value of 0.4 —the default value specified by the authors—.

2. *ProDrift*: trace-based detection, with a fixed-size window of 100 traces.

## 6.4 Results

The results show that *CRIER* performs better than *Martjushev et al.* and *ProDrift* in all datasets and for all metrics evaluated. In further detail:

- *Linear logs*. Table 4 shows the results for the linear gradual changes with slopes of 0.1%, 0.2%, 0.5%, and 1%. *CRIER* clearly outperforms the other proposals in terms of delay ($\Delta$), with values of an order of magnitude smaller than the obtained by the rest of approaches. *CRIER* achieves better results also in terms of average $F_{score}$, with *ProDrift* getting better values for some logs from the collection of linear logs with 0.1% slope, but severely underperforming in other logs —sw, OIR and ROI—, where no changes are detected by this approach. However, when changes happen at a faster rate —i.e. the slope of the linear combination increases—, *ProDrift* performance significantly drops, detecting almost no changes, while the results from *Martjushev et al.* improve, but still clearly worst than *CRIER*. Finally, in terms of $\Upsilon$, *CRIER* obtains better results than the rest of the algorithms in all cases but two.

Table 4: $F_{score}$, $\Delta$ and $\Upsilon$ values for the logs with linear gradual changes. The colors highlight the best performing approach for each metric.

|     |              | *CRIER* | | | *Martjushev et al.* | | | *ProDrift* | | |
|-----|--------------|---------|---------|---------|---------|---------|---------|---------|---------|---------|
|     |              | $F_{score}$ | $\Delta$ | $\Upsilon$ | $F_{score}$ | $\Delta$ | $\Upsilon$ | $F_{score}$ | $\Delta$ | $\Upsilon$ |
| cf  | slope = 0.1% | 0.8000 | 45.5000 | 79.86% | 0.6207 | 477.1111 | 27.90% | 0.8750 | 240.7143 | 66.28% |
|     | slope = 0.2% | 1.0000 | 14.4444 | 90.93% | 0.7826 | 100.4444 | 47.42% | 0.3636 | 240.0000 | 9.00% |
|     | slope = 0.5% | 1.0000 | 19.8889 | 78.67% | 0.7273 | 219.1250 | 18.00% | 0.0000 | - | 0.00% |
|     | slope = 1.0% | 1.0000 | 15.3333 | 66.22% | 0.3810 | 182.2500 | 10.11% | 0.0000 | - | 0.00% |
| cp  | slope = 0.1% | 0.7619 | 42.7500 | 79.23% | 0.7200 | 364.5556 | 29.44% | 0.8750 | 245.4286 | 64.96% |
|     | slope = 0.2% | 1.0000 | 24.7778 | 87.29% | 0.8000 | 137.7500 | 41.80% | 0.7143 | 235.7500 | 24.42% |
|     | slope = 0.5% | 1.0000 | 18.1111 | 77.94% | 0.7368 | 243.2857 | 30.44% | 0.0000 | - | 0.00% |
|     | slope = 1.0% | 1.0000 | 11.2222 | 70.33% | 0.4444 | 280.7500 | 8.89% | 0.0000 | - | 0.00% |
| pm  | slope = 0.1% | 0.3448 | 105.6000 | 43.72% | 0.5000 | 283.3333 | 15.09% | 0.8750 | 216.2857 | 68.73% |
|     | slope = 0.2% | 0.8000 | 38.3750 | 74.80% | 0.6154 | 139.7500 | 25.84% | 0.8235 | 203.1667 | 42.67% |
|     | slope = 0.5% | 0.9474 | 27.6667 | 78.28% | 0.5000 | 170.3333 | 29.78% | 0.3333 | 406.5000 | 22.00% |
|     | slope = 1.0% | 0.7368 | 15.4286 | 54.00% | 0.0000 | - | 0.00% | 0.5000 | 433.6667 | 22.44% |

Table 4: $F_{score}$, $\Delta$ and $\Upsilon$ values for the logs with linear gradual changes. The colors highlight the best performing approach for each metric. *(Continued)*

|  |  | CRIER | | | Martjushev et al. | | | ProDrift | | |
|---|---|---|---|---|---|---|---|---|---|---|
|  |  | $F_{score}$ | $\Delta$ | $\Upsilon$ | $F_{score}$ | $\Delta$ | $\Upsilon$ | $F_{score}$ | $\Delta$ | $\Upsilon$ |
| re | slope = 0.1% | 0.8571 | 27.6667 | 82.39% | 0.5806 | 337.7500 | 24.82% | 0.4615 | 288.3333 | 20.26% |
|  | slope = 0.2% | 1.0000 | 16.7778 | 88.20% | 0.7500 | 157.1111 | 34.47% | 0.0000 | - | 0.00% |
|  | slope = 0.5% | 1.0000 | 10.3333 | 82.39% | 0.4762 | 193.6000 | 16.78% | 0.0000 | - | 0.00% |
|  | slope = 1.0% | 1.0000 | 8.1111 | 68.78% | 0.1905 | 41.0000 | 12.00% | 0.0000 | - | 0.00% |
| rp | slope = 0.1% | 0.5385 | 60.5714 | 56.50% | 0.2000 | 881.0000 | 1.32% | 0.8750 | 236.1429 | 63.11% |
|  | slope = 0.2% | 0.9474 | 41.6667 | 81.89% | 0.0000 | - | 0.00% | 0.5000 | 243.3333 | 12.91% |
|  | slope = 0.5% | 0.9474 | 12.5556 | 71.94% | 0.0000 | - | 0.00% | 0.0000 | - | 0.00% |
|  | slope = 1.0% | 1.0000 | 12.0000 | 68.89% | 0.0000 | - | 0.00% | 0.0000 | - | 0.00% |
| sw | slope = 0.1% | 0.7273 | 54.0000 | 79.32% | 0.0000 | - | 0.00% | 0.0000 | - | 0.00% |
|  | slope = 0.2% | 1.0000 | 21.1111 | 86.33% | 0.0000 | - | 0.00% | 0.0000 | - | 0.00% |
|  | slope = 0.5% | 1.0000 | 13.3333 | 81.00% | 0.0000 | - | 0.00% | 0.0000 | - | 0.00% |
|  | slope = 1.0% | 1.0000 | 10.1111 | 66.44% | 0.0000 | - | 0.00% | 0.0000 | - | 0.00% |
| OIR | slope = 0.1% | 0.6087 | 45.4286 | 70.71% | 0.5455 | 474.1111 | 26.09% | 0.0000 | - | 0.00% |
|  | slope = 0.2% | 0.6364 | 14.7143 | 70.18% | 0.6667 | 173.1250 | 37.71% | 0.0000 | - | 0.00% |
|  | slope = 0.5% | 1.0000 | 8.2222 | 90.56% | 0.6957 | 186.1250 | 13.44% | 0.0000 | - | 0.00% |
|  | slope = 1.0% | 1.0000 | 11.6667 | 80.56% | 0.5600 | 60.8571 | 27.00% | 0.0000 | - | 0.00% |
| ORI | slope = 0.1% | 0.4444 | 81.5000 | 56.19% | 0.5000 | 435.8889 | 26.08% | 0.8750 | 325.1429 | 54.56% |
|  | slope = 0.2% | 0.7619 | 27.6250 | 77.40% | 0.5385 | 208.7143 | 22.62% | 0.3636 | 341.5000 | 3.36% |
|  | slope = 0.5% | 0.9474 | 34.4444 | 73.94% | 0.6087 | 209.1429 | 17.00% | 0.0000 | - | 0.00% |
|  | slope = 1.0% | 1.0000 | 11.4444 | 68.67% | 0.4348 | 31.4000 | 29.44% | 0.0000 | - | 0.00% |
| RIO | slope = 0.1% | 0.6364 | 80.2857 | 67.39% | 0.3636 | 209.5000 | 10.09% | 0.8750 | 216.5714 | 69.41% |
|  | slope = 0.2% | 1.0000 | 36.7778 | 83.76% | 0.2000 | 79.0000 | 7.58% | 0.9412 | 196.8571 | 56.84% |
|  | slope = 0.5% | 1.0000 | 27.6667 | 78.22% | 0.2000 | 32.0000 | 11.11% | 0.0000 | - | 0.00% |
|  | slope = 1.0% | 1.0000 | 17.2222 | 70.78% | 0.5000 | 145.0000 | 33.33% | 0.0000 | - | 0.00% |
| ROI | slope = 0.1% | 0.7619 | 31.3750 | 81.29% | 0.8421 | 287.0000 | 26.34% | 0.0000 | - | 0.00% |
|  | slope = 0.2% | 1.0000 | 13.7778 | 92.36% | 0.7368 | 194.7143 | 34.51% | 0.0000 | - | 0.00% |
|  | slope = 0.5% | 0.9000 | 9.0000 | 82.28% | 0.9412 | 247.3750 | 46.28% | 0.0000 | - | 0.00% |
|  | slope = 1.0% | 1.0000 | 7.3333 | 81.00% | 0.3333 | 223.6667 | 22.89% | 0.0000 | - | 0.00% |
| Average | slope = 0.1% | 0.6481 | 57.4677 | 69.66% | 0.4873 | 416.6944 | 17.70% | 0.5712 | 252.6599 | 37.89% |
|  | slope = 0.2% | 0.9146 | 25.0048 | 83.31% | 0.5090 | 148.8261 | 22.73% | 0.3706 | 243.4345 | 15.58% |
|  | slope = 0.5% | 0.9742 | 18.1222 | 79.52% | 0.4886 | 187.6234 | 18.31% | 0.0333 | 406.5000 | 2.22% |
|  | slope = 1.0% | 0.9737 | 11.9873 | 69.57% | 0.2844 | 137.8463 | 14.84% | 0.0500 | 433.6667 | 2.24% |

- *Gaussian logs.* Table 5 shows the results for the detection of gradual changes in logs that follow two Gaussian distributions. *CRIER* clearly outperforms *Martjushev et al.* in $F_{score}$, $\Delta$ and $\Upsilon$, obtaining values very close to 1 in $F_{score}$, delays always lower than 25 traces for almost all the logs, and $\Upsilon$ of more than 60 %.

Table 5: $F_{score}$, $\Delta$ and $\Upsilon$ values for the logs with Gaussian gradual changes. The colors highlight the best performing approach for each metric.

|  |  |  | CRIER | | | Martjushev et al. | | |
|---|---|---|---|---|---|---|---|---|
|  |  |  | $F_{score}$ | $\Delta$ | $\Upsilon$ | $F_{score}$ | $\Delta$ | $\Upsilon$ |
| cf | $\mu = 20$ | $\sigma^2 = 10$ | 0.9412 | 10.8750 | 61.22% | 0.2727 | 25.6667 | 20.92% |
|  | $\mu = 50$ | $\sigma^2 = 30$ | 1.0000 | 9.7778 | 70.78% | 0.5000 | 255.6000 | 6.84% |
| cp | $\mu = 20$ | $\sigma^2 = 10$ | 1.0000 | 8.2222 | 67.97% | 0.2105 | 170.0000 | 0.44% |
|  | $\mu = 50$ | $\sigma^2 = 30$ | 1.0000 | 8.6667 | 63.33% | 0.7368 | 283.0000 | 18.80% |
| pm | $\mu = 20$ | $\sigma^2 = 10$ | 1.0000 | 10.6667 | 55.56% | 0.3636 | 168.5000 | 22.22% |
|  | $\mu = 50$ | $\sigma^2 = 30$ | 1.0000 | 14.8889 | 66.36% | 0.6154 | 94.2500 | 44.44% |
| re | $\mu = 20$ | $\sigma^2 = 10$ | 1.0000 | 6.4444 | 63.40% | 0.2000 | 55.5000 | 12.42% |
|  | $\mu = 50$ | $\sigma^2 = 30$ | 1.0000 | 10.0000 | 62.94% | 0.4000 | 176.0000 | 9.25% |
| rp | $\mu = 20$ | $\sigma^2 = 10$ | 1.0000 | 7.1111 | 64.49% | 0.0000 | - | 0.00% |
|  | $\mu = 50$ | $\sigma^2 = 30$ | 1.0000 | 9.8889 | 60.22% | 0.0000 | - | 0.00% |
| sw | $\mu = 20$ | $\sigma^2 = 10$ | 1.0000 | 12.2222 | 63.83% | 0.0000 | - | 0.00% |
|  | $\mu = 50$ | $\sigma^2 = 30$ | 0.9474 | 8.1111 | 59.75% | 0.0000 | - | 0.00% |
| OIR | $\mu = 20$ | $\sigma^2 = 10$ | 1.0000 | 6.4444 | 64.05% | 0.5000 | 5.3333 | 61.44% |
|  | $\mu = 50$ | $\sigma^2 = 30$ | 1.0000 | 25.6667 | 61.93% | 0.5000 | 72.6667 | 26.34% |
| ORI | $\mu = 20$ | $\sigma^2 = 10$ | 1.0000 | 8.6667 | 62.31% | 0.2727 | 11.6667 | 29.85% |
|  | $\mu = 50$ | $\sigma^2 = 30$ | 1.0000 | 6.1429 | 47.40% | 0.6087 | 141.7143 | 20.44% |
| RIO | $\mu = 20$ | $\sigma^2 = 10$ | 0.9412 | 10.3750 | 60.57% | 0.0000 | - | 0.00% |
|  | $\mu = 50$ | $\sigma^2 = 30$ | 1.0000 | 15.4444 | 64.34% | 0.3636 | 13.0000 | 21.13% |
| ROI | $\mu = 20$ | $\sigma^2 = 10$ | 1.0000 | 6.7778 | 69.28% | 0.4444 | 253.5000 | 17.86% |
|  | $\mu = 50$ | $\sigma^2 = 30$ | 0.9000 | 7.3333 | 68.07% | 0.8421 | 254.5000 | 36.52% |
| Average | $\mu = 20$ | $\sigma^2 = 10$ | 0.9882 | 8.7806 | 63.50% | 0.2264 | 98.5952 | 16.03% |
|  | $\mu = 50$ | $\sigma^2 = 30$ | 0.9847 | 11.5921 | 61.59% | 0.4567 | 161.3414 | 19.66% |

- *Exponential logs.* Table 6 shows the results for the detection of gradual changes in logs that follow three exponential distributions with $\lambda$ set to 0.05, 0.1, and 0.5. *CRIER* also obtains the best results, with $\Delta$ values two orders of magnitude smaller than those of *Martjushev et al.*. In terms of $F_{score}$ and $\Upsilon$, *CRIER* still achieves results that are, on average, almost twice better than *Martjushev et al.*. It is worth noting that $\Upsilon$ values are usually lower with the exponential distribution, if compared with the other distributions.

Table 6: $F_{score}$, $\Delta$ and $\Upsilon$ values for the logs with exponential gradual changes. The colors highlight the best performing approach for each metric.

|  |  | CRIER | | | Martjushev et al. | | |
|---|---|---|---|---|---|---|---|
|  |  | $F_{score}$ | $\Delta$ | $\Upsilon$ | $F_{score}$ | $\Delta$ | $\Upsilon$ |
| cf | $\lambda = 0.05$ | 0.5000 | 6.0000 | 23.02% | 0.3333 | 205.2500 | 11.99% |
|  | $\lambda = 0.10$ | 1.0000 | 5.2222 | 60.00% | 0.2857 | 38.6667 | 11.43% |
|  | $\lambda = 0.50$ | 0.3636 | 2.0000 | 19.05% | 0.2857 | 9.0000 | 33.33% |
| cp | $\lambda = 0.05$ | 0.4615 | 8.8333 | 12.39% | 0.5000 | 297.8000 | 15.83% |
|  | $\lambda = 0.10$ | 0.9412 | 6.3750 | 53.81% | 0.2222 | 300.5000 | 6.51% |
|  | $\lambda = 0.50$ | 0.3636 | 3.0000 | 17.46% | 0.1111 | 285.0000 | 0.79% |

*Continues on next page*

Table 6: $F_{score}$, $\Delta$ and $\Upsilon$ values for the logs with exponential gradual changes. The colors highlight the best performing approach for each metric. *(Continued)*

|  |  | *CRIER* |  |  | *Martjushev et al.* |  |  |
|---|---|---|---|---|---|---|---|
|  |  | $F_{score}$ | $\Delta$ | $\Upsilon$ | $F_{score}$ | $\Delta$ | $\Upsilon$ |
| pm | $\lambda = 0.05$ | 0.4167 | 9.8000 | 25.82% | 0.2000 | 166.0000 | 11.11% |
|  | $\lambda = 0.10$ | 1.0000 | 8.7778 | 49.68% | 0.2000 | 217.0000 | 11.11% |
|  | $\lambda = 0.50$ | 0.2000 | 3.0000 | 8.73% | 0.0000 | - | 0.00% |
| re | $\lambda = 0.05$ | 0.6364 | 7.4286 | 34.69% | 0.5000 | 92.6667 | 22.14% |
|  | $\lambda = 0.10$ | 1.0000 | 3.7778 | 57.46% | 0.4348 | 47.6000 | 17.14% |
|  | $\lambda = 0.50$ | 0.7143 | 2.8000 | 44.44% | 0.1905 | 7.5000 | 11.11% |
| rp | $\lambda = 0.05$ | 0.4800 | 8.3333 | 17.43% | 0.0000 | - | 0.00% |
|  | $\lambda = 0.10$ | 0.9412 | 4.2500 | 52.70% | 0.0000 | - | 0.00% |
|  | $\lambda = 0.50$ | 0.0000 | - | 0.00% | 0.0000 | - | 0.00% |
| sw | $\lambda = 0.05$ | 0.2963 | 11.5000 | 15.51% | 0.0000 | - | 0.00% |
|  | $\lambda = 0.10$ | 0.7000 | 5.7143 | 38.73% | 0.0000 | - | 0.00% |
|  | $\lambda = 0.50$ | 0.8000 | 3.3333 | 50.79% | 0.0000 | - | 0.00% |
| OIR | $\lambda = 0.05$ | 1.0000 | 5.2222 | 69.06% | 0.6154 | 50.3750 | 41.25% |
|  | $\lambda = 0.10$ | 1.0000 | 3.7778 | 60.48% | 0.6667 | 20.8889 | 61.27% |
|  | $\lambda = 0.50$ | 0.6154 | 2.5000 | 36.51% | 0.6667 | 13.1111 | 99.21% |
| ORI | $\lambda = 0.05$ | 0.5000 | 9.8333 | 29.42% | 0.5600 | 52.0000 | 29.58% |
|  | $\lambda = 0.10$ | 0.6000 | 6.5000 | 35.40% | 0.5000 | 22.1429 | 53.02% |
|  | $\lambda = 0.50$ | 0.3636 | 2.5000 | 18.25% | 0.4167 | 11.2000 | 49.21% |
| RIO | $\lambda = 0.05$ | 0.8000 | 12.1250 | 46.44% | 0.2000 | 160.0000 | 11.11% |
|  | $\lambda = 0.10$ | 0.9412 | 8.1250 | 56.67% | 0.2000 | 233.0000 | 11.11% |
|  | $\lambda = 0.50$ | 0.3636 | 3.5000 | 16.67% | 0.2000 | 192.0000 | 11.11% |
| ROI | $\lambda = 0.05$ | 0.5000 | 6.6667 | 29.26% | 0.7368 | 247.1429 | 25.34% |
|  | $\lambda = 0.10$ | 0.8750 | 4.4286 | 46.98% | 0.5263 | 236.6000 | 38.10% |
|  | $\lambda = 0.50$ | 0.8000 | 2.8333 | 53.17% | 0.2222 | 251.5000 | 22.22% |
| *Average* | $\lambda = 0.05$ | 0.5591 | 8.5742 | 31.11% | 0.3646 | 158.9043 | 17.37% |
|  | $\lambda = 0.10$ | 0.8999 | 5.6948 | 50.21% | 0.3036 | 139.5498 | 22.03% |
|  | $\lambda = 0.50$ | 0.4584 | 2.8296 | 27.34% | 0.2093 | 109.9016 | 21.52% |

- *Constant logs*. Table 7 shows the results for the detection of gradual changes in logs that follow three different constant distributions. In these logs, *CRIER* obtained the best results, with a perfect $F_{score}$ for all the cases, average $\Delta$ of less than 10 traces, and $\Upsilon$ values above 80 %. On the contrary, *Martjushev et al.* achieves average $\Delta$ values between 140 and 220 traces, $\Upsilon$ values always below 60 %, and average $F_{score}$ values that do not reach 0.6, with several cases in which no change is detected —as rp and sw.

Table 7: $F_{score}$, $\Delta$ and $\Upsilon$ values for the logs with constant gradual changes. The colors highlight the best performing approach for each metric.

|  |  |  | CRIER | | | Martjushev et al. | | |
|---|---|---|---|---|---|---|---|---|
|  |  |  | $F_{score}$ | $\Delta$ | $\Upsilon$ | $F_{score}$ | $\Delta$ | $\Upsilon$ |
|  | $p = 0.5$ | $n = 100$ | 1.0000 | 5.4444 | 84.33% | 0.3158 | 230.0000 | 2.22% |
| cf | $p = 0.5$ | $n = 200$ | 1.0000 | 6.2222 | 93.00% | 0.7273 | 172.5000 | 26.06% |
|  | $p = 0.5$ | $n = 500$ | 1.0000 | 6.2222 | 97.36% | 0.9474 | 127.2222 | 46.69% |
|  | $p = 0.5$ | $n = 100$ | 1.0000 | 5.6667 | 84.33% | 1.0000 | 263.5556 | 30.44% |
| cp | $p = 0.5$ | $n = 200$ | 1.0000 | 5.1111 | 92.22% | 0.8889 | 287.7500 | 28.72% |
|  | $p = 0.5$ | $n = 500$ | 1.0000 | 4.8889 | 96.56% | 1.0000 | 125.7778 | 53.64% |
|  | $p = 0.5$ | $n = 100$ | 1.0000 | 4.5556 | 86.44% | 0.0000 | - | 0.00% |
| pm | $p = 0.5$ | $n = 200$ | 1.0000 | 4.4444 | 94.72% | 0.0000 | - | 0.00% |
|  | $p = 0.5$ | $n = 500$ | 1.0000 | 6.6667 | 97.67% | 0.6154 | 81.7500 | 37.73% |
|  | $p = 0.5$ | $n = 100$ | 1.0000 | 3.2222 | 84.22% | 0.1053 | 93.0000 | 0.78% |
| re | $p = 0.5$ | $n = 200$ | 1.0000 | 3.5556 | 93.39% | 0.5714 | 230.1667 | 17.33% |
|  | $p = 0.5$ | $n = 500$ | 1.0000 | 3.0000 | 96.49% | 0.7273 | 213.7500 | 37.71% |
|  | $p = 0.5$ | $n = 100$ | 1.0000 | 17.1111 | 63.11% | 0.0000 | - | 0.00% |
| rp | $p = 0.5$ | $n = 200$ | 1.0000 | 5.2222 | 90.94% | 0.0000 | - | 0.00% |
|  | $p = 0.5$ | $n = 500$ | 1.0000 | 5.8889 | 96.89% | 0.0000 | - | 0.00% |
|  | $p = 0.5$ | $n = 100$ | 1.0000 | 4.6667 | 87.33% | 0.0000 | - | 0.00% |
| sw | $p = 0.5$ | $n = 200$ | 1.0000 | 4.2222 | 94.44% | 0.0000 | - | 0.00% |
|  | $p = 0.5$ | $n = 500$ | 1.0000 | 6.8889 | 96.29% | 0.0000 | - | 0.00% |
|  | $p = 0.5$ | $n = 100$ | 1.0000 | 2.7778 | 89.22% | 0.6154 | 39.7500 | 49.44% |
| OIR | $p = 0.5$ | $n = 200$ | 1.0000 | 3.0000 | 95.06% | 0.5455 | 191.8333 | 19.22% |
|  | $p = 0.5$ | $n = 500$ | 1.0000 | 3.1111 | 96.73% | 0.8571 | 198.7778 | 38.87% |
|  | $p = 0.5$ | $n = 100$ | 1.0000 | 5.4444 | 86.56% | 0.4545 | 88.6000 | 25.89% |
| ORI | $p = 0.5$ | $n = 200$ | 1.0000 | 3.5556 | 96.06% | 0.8571 | 232.5556 | 19.61% |
|  | $p = 0.5$ | $n = 500$ | 1.0000 | 3.8889 | 97.24% | 0.6667 | 256.4444 | 40.60% |
|  | $p = 0.5$ | $n = 100$ | 1.0000 | 12.1111 | 83.67% | 0.0000 | - | 0.00% |
| RIO | $p = 0.5$ | $n = 200$ | 1.0000 | 10.0000 | 93.11% | 0.5000 | 171.6667 | 33.06% |
|  | $p = 0.5$ | $n = 500$ | 1.0000 | 9.8889 | 97.40% | 0.3636 | 21.0000 | 20.22% |
|  | $p = 0.5$ | $n = 100$ | 1.0000 | 3.3333 | 90.89% | 0.8421 | 245.5714 | 42.22% |
| ROI | $p = 0.5$ | $n = 200$ | 1.0000 | 3.0000 | 96.33% | 1.0000 | 254.4444 | 44.94% |
|  | $p = 0.5$ | $n = 500$ | 1.0000 | 3.3333 | 98.36% | 0.9474 | 153.3333 | 54.98% |
|  | $p = 0.5$ | $n = 100$ | 1.0000 | 6.4333 | 83.98% | 0.3333 | 160.0795 | 16.53% |
| Average | $p = 0.5$ | $n = 200$ | 1.0000 | 4.8333 | 94.03% | 0.5090 | 220.1310 | 18.10% |
|  | $p = 0.5$ | $n = 500$ | 1.0000 | 5.3778 | 97.07% | 0.6125 | 147.2569 | 33.04% |

Table 8 shows the findings of the BAIN test performed over the results obtained by all approaches. In this test, the hypothesis $H_1$ considers that *CRIER* achieves better results than the other *state-of-the-art* approaches —i.e., *CRIER > Martjushev et al.* and *CRIER > ProDrift*—, while $H_1^c$ considers the contrary hypothesis. As Table 8 shows, the Bayes factor $BF_{H_1 H_1^c}$ for all metrics in all distributions is greater than 100, meaning that there is an extreme evidence for $H_1$, i.e., *CRIER* is better than *ProDrift* and *Martjushev et al.*.

## 6.5 Discussion

As shown in the previous Section, *CRIER* clearly outperforms the current *state-of-the-art* approaches. In particular, for linear logs, *ProDrift* has a higher accuracy for some logs —cf, cp, rp, ORI and RIO with

Table 8: BAIN test results for the validation logs, where $BF_{H_i H_i^c}$ shows the Bayesian factor for $H_i$ against its complementary $H_i^c = \neg H_i$. The most likely hypothesis is shown shaded in blue.

|  |  | $BF_{H_i H_i^c}$ | | |
| --- | --- | --- | --- | --- |
|  |  | $F_{score}$ | $\Delta$ | $\Upsilon$ |
| Linear logs | $H_1$: CRIER > Martjushev et al. and CRIER > ProDrift | $1.20 \times 10^{12}$ | $1.99 \times 10^{13}$ | $2.28 \times 10^{13}$ |
|  | $H_2$: CRIER < Martjushev et al. | $4.30 \times 10^{-12}$ | $3.40 \times 10^{-16}$ | $7.42 \times 10^{-48}$ |
|  | $H_3$: CRIER < ProDrift | $9.56 \times 10^{-23}$ | $1.96 \times 10^{-14}$ | $8.96 \times 10^{-55}$ |
|  | $H_4$: CRIER = Martjushev et al. | $5.81 \times 10^{-10}$ | $5.10 \times 10^{-14}$ | $2.10 \times 10^{-45}$ |
|  | $H_5$: CRIER = ProDrift | $1.83 \times 10^{-20}$ | $2.16 \times 10^{-12}$ | $2.71 \times 10^{-52}$ |
| Gaussian logs | $H_1$: CRIER > Martjushev et al. | $2.29 \times 10^{13}$ | $4.74 \times 10^{7}$ | $2.29 \times 10^{13}$ |
|  | $H_2$: CRIER < Martjushev et al. | $8.57 \times 10^{-29}$ | $2.11 \times 10^{-8}$ | $2.04 \times 10^{-31}$ |
|  | $H_3$: CRIER = Martjushev et al. | $1.52 \times 10^{-26}$ | $1.75 \times 10^{-6}$ | $3.79 \times 10^{-29}$ |
| Exponential logs | $H_1$: CRIER > Martjushev et al. | $1.16 \times 10^{7}$ | $3.05 \times 10^{10}$ | $6.60 \times 10^{2}$ |
|  | $H_2$: CRIER < Martjushev et al. | $8.60 \times 10^{-8}$ | $3.28 \times 10^{-11}$ | $1.51 \times 10^{-3}$ |
|  | $H_3$: CRIER = Martjushev et al. | $9.03 \times 10^{-6}$ | $3.94 \times 10^{-9}$ | $0.10$ |
| Constant logs | $H_1$: CRIER > Martjushev et al. | $9.00 \times 10^{12}$ | $2.29 \times 10^{13}$ | $2.29 \times 10^{13}$ |
|  | $H_2$: CRIER < Martjushev et al. | $1.21 \times 10^{-13}$ | $1.38 \times 10^{-33}$ | $5.64 \times 10^{-75}$ |
|  | $H_3$: CRIER = Martjushev et al. | $1.75 \times 10^{-11}$ | $2.94 \times 10^{-31}$ | $2.01 \times 10^{-72}$ |

*slope* of 0.1% and pm with *slopes* of 0.1% and 0.2%— in which the traces of the new behavior are added at a slower rate at the beginning of the change, while *CRIER* has clearly the best performance with higher insertion rates. This low frequency forces the conformance checking metrics to change slowly, as the behavior causing the change appears rarely, which in turn leds the slope of the regression to stay invariant until the behaviour is more frequent; resulting in a late/early confirmation of the change depending on whether we are at the beginning or at the end of the change. Consequently, *CRIER* obtains high delays in detecting the start of the change, and premature detections at the end of the change, having a high number of false positives that lead to a decrease in accuracy. Particularly interesting are the logs with slowest changes —0.1% slope, i.e., 1,000-trace change regions—, where *ProDrift* obtains the best $F_{score}$ in 6 out of 10 logs. For these 6 logs, *CRIER* detects a lower change region overlapping, but obtains the best results in terms of $\Delta$, which reinforces the idea that the detection errors are due to false positives in determining the end of the change. However, with respect to the average values of the metrics, *CRIER* is the best positioned since *ProDrift* is not able to detect any change in 3 of the logs, while *Martjushev et al.* obtains many false positives, detecting less than 18% of the traces as a region of change.

In the remaining linear logs, the results of *ProDrift* suffer a significant degradation, detecting changes only in 2 of 20 logs with a greater slope than 0.1 %, in which it obtains $F_{score}$ values of 0.5 or less, and $\Delta$ of more than 400 traces. This degradation may be due to the fact that, as changes are much faster, the algorithm does not have enough information to correctly extract the distribution of *partially-ordered-runs* before and after the change, not being able to correctly detect the drifts.

In logs with Gaussian distributions, the results obtained by *CRIER* are clearly better than those of *Martjushev et al.*, with $F_{score}$ of 1.0 in 17 of 20 logs. These logs are characterized by relatively fast changes, but with both a slow start and termination. The results are consistent with what was previously established for the linear changes: as these are relatively fast changes —slightly more than 150 traces in the slowest case— the slope of the regression changes abruptly, facilitating the detection of changes. The values of Δ and ϒ obtained by *Martjushev et al.* are particularly indicative. The slow start and termination of the new behavior lead to smaller changes in the distributions of the features, making the statistical test employed by this approach unable to detect the drifts and causing the beginning of the changes to be detected late —increasing Δ values— while their termination is detected prematurely —resulting in low ϒ values—.

Something similar happens for logs with constant distribution. In this case, *CRIER* obtains the best results, with a perfect average $F_{score}$ and an average ϒ of around 90 %. This is because, as soon as the change starts, half of the observed behavior is part of the new model. Consequently, conformance metrics change very fast, which means that the slope of the regression is significantly modified and changes are detected almost instantly. For *Martjushev et al.*, there are several logs in which no change is detected —`rp` and `sw` for all cumulative probability functions—. The poor performance of this algorithm for the same change patterns may be caused by the selected features since they are not able to capture such changes. Perhaps selecting other features that might be more sensitive to this type of changes would improve the results, however a more in-depth analysis should be performed to confirm this intuition.

Finally, for logs with exponential changes, the results of *CRIER* are quite degraded when compared to the other distributions. These results are due to two distinct situations: on the one hand, in logs with $\lambda = 0.5$, changes are so fast —only 14 traces— that *CRIER* is not able to detect them as gradual, leading to false negatives; and, on the other hand, in logs with $\lambda = 0.05$, changes have such long tails that they are terminated prematurely, and consequently, a high number of false positives are detected in traces that belong to the actual change but have not been detected as part of the drift. This behavior of the algorithm is reflected by the values of Δ and ϒ. As Table 6 shows, all changes are detected very close where they begin —with Δ below 10 traces in almost all cases—, but ϒ values remain around 20%, indicating that only the initial part of the change region is detected. In addition, logs in which *CRIER* obtains the lowest ϒ coincide with those with the worst $F_{score}$, confirming this hypothesis.

# 7 Conclusions and Future Work

In this paper, we presented *CRIER*, an offline gradual-drift detection algorithm. *CRIER* is based on the hypothesis that change detection and classification can be addressed by analyzing how the fitness and precision of the process models —the old model and the new one after the change— vary from the beginning to the end of the gradual change. Specifically, the hypothesis is that at the beginning there is a modification of the fitness, keeping the precision; while at the end the precision changes. This hypothesis has been mathematically demonstrated.

The approach has been validated using a synthetic dataset with 120 logs that present different change patterns and gradual change distributions, from linear to Gaussian, exponential and constant. The experiments show that *CRIER* outperforms the results of the main *state-of-the-art* approaches in terms of accuracy ($F_{score}$); delay (Δ), so drifts are detected faster; and change region overlapping (ϒ), so the time interval during which two processes coexist is clearly identified.

Furthermore, *CRIER* is more robust for all types of change distributions and particularly better for those in which the traces of the new model start to arise with a higher frequency at the beginning of the gradual drift, such as Gaussian, exponential, and constant distributions. Nonetheless, our approach has proven to be more consistent even when the former conditions are not present such as when the drift follows a linear distribution with a low slope, being able to detect all the change patterns with less delay $\Delta$ and better $\Upsilon$.